\documentclass[review]{elsarticle}
\usepackage{amsmath,amsthm, amssymb}
\usepackage{graphicx}

\usepackage{algorithmic,algorithm}

\usepackage{subfigure}

\newtheorem{MyThe}{Theorem}
\newtheorem{MyCor}{Corollary}
\newtheorem{MyLem}{Lemma}
\newtheorem{MyDef}{Definition}
\begin{document}
\begin{frontmatter}
\title{MSTGD:A Memory Stochastic sTratified Gradient Descend Method with an Exponential Convergence Rate\tnoteref{mytitlenote}}
\tnotetext[mytitlenote]{Research supported by Chinese National Office for Philosophy and Social Sciences(19FTJB003),National Bureau of Statistics of China($2016LY64$), Natural Science Foundation of Guangdong Province (10451032001006140).}
\author[mymainaddress,mymainaddress2]{Aixiang(Andy) Chen}
\author[mymainaddress3]{Jinting Zhang}
\author[mymainaddress,mymainaddress2]{Zanbo Zhang}
\author[mymainaddress2]{Zhihong Li}
\address[mymainaddress]{Institute of Artificial Intelligence and Deep Learning, Guangdong University of Finance and Economics,\small Guangzhou 510320,China}
\address[mymainaddress2]{\footnotesize School of Statistics and Mathematics, Guangdong University of Finance and Economics, Guangzhou 510320, China}
\address[mymainaddress3]{\footnotesize Department of Statistics and Data Science, National University of Singapore,  Singapore 117546}
\begin{abstract}
The fluctuation effect of gradient expectation and variance caused by parameter update between consecutive iterations is neglected or confusing by current mainstream gradient optimization algorithms.Using this fluctuation effect, combined with the stratified sampling strategy, this paper designs a novel \underline{M}emory \underline{S}tochastic s\underline{T}ratified Gradient Descend(\underline{MST}GD) algorithm with an exponential convergence rate. Specifically, MSTGD uses two strategies for variance reduction: the first strategy is to perform variance reduction according to the proportion p of used historical gradient, which is estimated from the mean and variance of sample gradients before and after iteration, and the other strategy is stratified sampling by category. The statistic \ $\bar{G}_{mst}$\  designed under these two strategies can be adaptively unbiased, and its variance decays at a geometric rate. This enables  MSTGD  based on $\bar{G}_{mst}$ to obtain an exponential convergence rate of the form $\lambda^{2(k-k_0)}$($\lambda\in (0,1)$,k is the number of iteration steps,$\lambda$ is a variable related to  proportion p).Unlike most other algorithms that claim to achieve an exponential convergence rate, the convergence rate is independent of parameters such as dataset size N, batch size n, etc., and can be achieved at a constant step size.Theoretical and experimental results show the effectiveness of MSTGD
\end{abstract}


\end{frontmatter}
\section{Introduction}\label{intro}
In this paper, we consider the problem of optimizing the objective function of the form \ \ref{eq:cost1}\  given a training set (X, Y) with N records.
\begin{equation}\label{eq:cost1}
\arg\min_{W\in R^\wp}J(W)=\frac{1}{N}\sum_{i=1}^{N}J(W;x^{(i)},y^{(i)})=\frac{1}{N}\sum_{i=1}^{N}J_{i},
\end{equation}
where $W=[w_1,w_2,\cdots,w_{\wp}]^T$  is the optimization parameter, which can be a parameter set composed of connection weights in a deep network model. In different contexts, $W$ can refer to model parameters or it can represent a deep network model.

The objective function represented by Equation\ \ref{eq:cost1}\  p is in the form of a finite sum of the training sample set, and the term \ $J_i=J(W;x^{(i)},y^{(i)})$\  in the summation is the loss function for sample\ $(x^{(i)},y^{(i)})$\ .The lost function can be least squares, cross entropy, etc.

Problem of the form \ref{eq:cost1} are of broad interest, as they encompass a variety of problems in statistics,machine learning and optimization. For example, many problems such as image processing and visual recognition\cite{andy2020,LeCun1998-3,Ciresan2010,Ciresan2012,KrizhevskySH12-2}, speech recognition\cite{Graves2006,Graves2013-4,Shillingford2018}, machine translation and natural language understanding\cite{Yonghui2016-4} can be reduced to the optimization problem in the form of Equation\ \ref{eq:cost1}\ .

Because of their wide applicability, it is important to carefully design and develop more efficient solver to such problems. Gradient descent method (GDM) which originated from the work of Robbins and Monro et al.\cite{Robbins1951}  and its series of improved algorithms are the current mainstream effective optimization algorithms for such problems. However, Achieving the linear convergence rate of the full gradient while keeping a low iteration cost as the stochastic gradient descend is still an open and challenging problem.

This paper proposes a new strategy called \underline{M}emory \underline{S}tochastic s\underline{T}ratified Gradient Descend(\underline{MST}GD), According to which a statistic called \ $\bar{G}_{mst}$\ is designed,whose subscripts come from the first three letters of MSTGD. MSTGD iterates as follow
\begin{equation}\label{eq:weightupdate}
W_{k+1}=W_{k}-h\bar{G}^{k}_{mst}(W_k)=W_{k}-h\sum\limits_{j=1}^{C}\textit{w}_j\cdot G_{j}^{k}(W_k),
\end{equation}
where $k$ represents the index of the iteration, $h$ is the step size of the algorithm, and the italicized\ $\textit{w}_j$\ represents the class weight, which can be determined according to the ratio of the total number of \ $j^{th}(j\in\{1,2,\cdots,C\})$\ class samples to the total number of samples.

For the sake of brevity, this paper will\ $\bar{G}^{k}_{mst}(W_k),G_{j}^{k}(W_k)$\ abbreviated as\ $\bar{G}^{k}_{mst},G_{j}^{k}$\ respectively.p

The value of $\bar{G}_{mst}^k$ in formulae (\ref{eq:weightupdate})  is obtained by averaging a $C$-dimensions auxiliary storage vector $G$ by components. The auxiliary vector $G$ tracks the gradient signal ever used,
and this is why MSTGD being named as the memory algorithm.
In the $k$-th iteration we set
\begin{equation}\label{eq:Gvector}
G_{j}^{k}=
\begin{aligned}
p_j^k\cdot G_{j}^{k-1}+q_j^k\cdot g(W_k,\xi_j^k) &\ for\ each\ \substack{j\in \{1,\cdots,C\},\\k\in \{1,2,\cdots,\}},
\end{aligned}
\end{equation}
where $\xi_j^k$ is a random index of sample in category $j$,the $g(W_k,\xi_j^k)$ is random gradient generated by network $W_k$ inputting this sample. The stochastic gradient\ $g(W_k,\xi_j^k)$\  generated by the independent sample\ $\xi_j^k$\ ensures that it is independent of the historical gradient\ $G_{j}^{k-1}$\ .

When $p_j^k=0,q_j^k=1$, Equation (\ref{eq:Gvector}) is simplified into the memoryless form of\ $G_j^k=g(W_k,\xi_j^k)$\ . In this case, the mean value of traditional stratified sampling\ $\bar{G}_{st}^k$\ is calculated according to\ $\sum\limits_{j=1}^C\omega_j\cdot G_j^k$\  in Equation (\ref{eq:weightupdate}) . The subscript of\ $\bar{G}_{st}^k$\ is the first two letters of the word stratification.

This paper has made the following three main contributions:
\begin{itemize}
\item A novel  statistic\ $\bar{G}_{mst}$\  is proposed,and gives the \ $p,q$\ condition that makes \ $\bar{G}_{mst}$\ adaptively unbiased(\ $p_j^k,q_j^k$\ in Formulae (\ref{eq:valuepq}) or (\ref{The:valuepq})).
\item A new strategy called p-based variance reduction is proposed, which enables the variance of\ $\bar{G}_{mst}$\ to decay at a geometric rate.Prior to this, it was generally believed that in order to exponentially decay the variance of the mini-batch stochastic gradient, it was necessary to use a dynamic sample size strategy such as the exponential growth of the sample size.
\item This paper proves that MSTGD can achieve exponential convergence rate in the form of \ $\lambda^{2(k-k_0)}$($0<\lambda<1$, k\ is the number of iteration steps)under the condition of constant step size h and constant sample size.
\end{itemize}

\subsection{Related works}
The gradient descent method(GDM) originated from the work of Robbins and Monro et al.\cite{Robbins1951} and its improved algorithms are the current mainstream optimization algorithms. GDM iterates as the form
\begin{equation}\label{eq:sgd-likeupdate}
W_{k+1}=W_{k}-h\cdot\bar{g}=W_{k}-\frac{h}{n}\sum\limits_{i=1}^{n}g(W_k,i),
\end{equation}
where $\bar{g}$ is sample gradient mean,$g(W_k,i)$ is gradient of sample i. The sample size is 1 or N (population size), the corresponding algorithms are called \underline{S}tochastic \underline{G}radient \underline{D}escent(SGD) and \underline{F}ull \underline{G}radient \underline{D}escent(FGD),respectively. If $1<n<N$, the corresponding algorithm is called the mini-\underline{Batch} stochastic gradient descent algorithm(Batch).In this paper, SGD, FGD and Batch are collectively referred to as GDM, whicph belongs to the category of SGD-like Algorithms

As a kind of SGD-like Algorithms, one of the main features of MSTGD is to determine the next moving direction by weighting the current gradient and the historical gradient stored in the auxiliary vector G. The weight coefficient of the historical gradient plays a role in variance reduction while ensuring unbiased. Similar to MSTGD, other SGD-lik algorithms that use historical gradient information  include Momentum\cite{Qian1999} and its improved algorithms\cite{Nesterov,Kingma2015,Dozat2016}, Adam and AdaMax\cite{Kingma2015}, Nadam\cite{Dozat2016}, NAG\cite{Nesterov}, AMSGrad\cite{Sashank2018}. As far as we know, these algorithms are lacking in the discussion of the influence of the weight coefficient of the historical gradient on the unbiasedness and variance of the gradient direction. This paper will clarify how the weight coefficient affects the unbiasedness of the statistic $\bar{G}_{mst}$, and give sufficient conditions to exponentially decay the variance of $\bar{G}_{mst}$.It is theoretically proved that the algorithm MSTGD designed in this paper has a linear convergence rate that Momentum, Adam, AdaMax, Nadam, NAG, AMSGrad etc. have not claim.

The second feature of MSTGD is that when the objective function satisfies the assumption of continuous strong convexity, the theoretical linear convergence rate can be achieved under a constant step size. There is a lot of work to improve the algorithm around the step size. For example, most of the step size adaptive algorithms Adagrad\cite{Duchi2011},Adadelta\cite{Matthew2012},Adam and Adamax\cite{Kingma2015},Nadam\cite{Dozat2016}, AMSGrad\cite{Sashank2018}, etc., use the gradient second-order moment information to set the dynamic step size. To the best of the authors' knowledge, there is little discussion of whether these improvements can improve the convergence order of the algorithm.

The third feature of MSTGD is to use a more efficient stratified sampling strategy to ensure a linear convergence rate without consuming too much storage space. Work on improving the performance of stochastic gradient optimization algorithms through more efficient stratified sampling strategies originally came from an algorithm named SSAG\cite{Andy2018}.The calculation of the gradient direction $\bar{G}_{st}$ of SSAG also uses an auxiliary vector G similar to this paper, but $\bar{G}_{st}$ does not involve the preservation of historical gradients when updating the components in G,which is an essential difference with the memory-type stochastic stratified gradient $\bar{G}_{mst}$ in this paper. As far as we know, the algorithms that claim to achieve linear convergence rate include SAG\cite{Roux2012},SAGA\cite{Defazio2014},SVRG\cite{Johnson2013},DSSM(Dynamic Sample Size Methods)\cite{Hashemi2015},SSAG\cite{Andy2018}, etc., but the linear convergence rate of these algorithms often needs to consume too much memory. For example, SAG must record the gradient for each sample, which is a large storage overhead in the case of massive training data. In contrast, MSTGD only needs to save gradient  for each category of data, requiring much less storage space.For another example, in order to achieve a linear convergence rate, DSSM adopts a dynamic sample capacity strategy of exponential growth in the form of $\lceil\tau^{k-1}\rceil$($\tau>1$). Correspondingly, the computing resources required for a single iteration also increase exponentially, while the storage resources and computing resources required for each iteration of MSTGD  are constant.


\subsection{Notation and basic definition}
The following is notation and basic definition used in this paper.
\begin{itemize}
\item $\xi_j^k:N_j\rightarrow n_j^k$ indicates that $n_j^k$ samples are drawn from the $j^{th}$ subpopulation at the $k^{th}$ iteration.$f=\frac{n}{N},f_j^k=\frac{n_j^k}{N_j}$ is the corresponding sampling ratio. $n^k_j=1$ by default.
\item $g(W_k,\xi_j^k):$ the stochastic gradient generated when a sample is randomly drawn from the $j^{th}$ subpopulation and fed into the network $W_k$. $g(W_k,i):$ the gradient generated when the $i^{th}$ sample in the set is input into the network $W_k$
\item $\nabla J(W_k)=\nabla J_N(W_k)=\frac{1}{N}\sum\limits_{i=1}^Ng(W_k,i)$:population mean of gradient at iteration k. $\nabla J_n(W_k)=\bar{g}=\frac{1}{n}\sum\limits_{i=1}^ng(W_k,i)$: sample mean of gradient at iteration k.
\item $\sigma_k^2=\frac{1}{N}\sum\limits_{i=1}^N(g(W_k,i)-\nabla J_N(W_k))^2$:population variance of gradient at iteration k.
\item $V(\cdot)$:the variance of the variable in parentheses.
\item $s_k^2:=V(\bar{g})=\frac{(1-f_k)\sigma_k^2}{n_k}$:the variance of the sample gradient mean at the $k^{th}$ iteration
\end{itemize}

\section{Properties of the statistic \ $\bar{G}_{mst}$\ }
\subsection{Conditions on $p,q$ to ensure the unbiasedness of $\bar{G}_{mst}$}
The basic idea to ensure $\bar{G}_{mst}$  an unbiased estimation is
to transfer the gradient signal of the previous iteration to the current iteration
in an appropriate proportion.
This result is given in the below theorem.
\begin{MyThe}\label{the:unbias}
Let $g(W_{k-1},\xi_j^{k-1}),g(W_{k},\xi_j^k)$, $j=1,\cdots,C$ denote the random gradients of the networks $W_{k-1},W_{k}$
with  random sample $\xi_j^{k-1},\xi_j^k$ from the $j$-th class as input
(that is, the random gradients are produced in two consecutive iterations),
and $E^{k-1}_j,E^k_j$ be their expectations respectively.
If $\frac{p_j^k}{1-q_j^k}=\frac{E^k_j}{E^{k-1}_j}$, then $\bar{G}_{mst}^k$ is an unbiased estimation of the population mean $\bar{G}$,
i.e., $E(\bar{G}_{mst}^k)=\bar{G}^k$, or $E(\bar{G}_{mst})=\bar{G}$ for simplification.
\end{MyThe}

In Theorem \ref{the:unbias}, the expectations $E^{k-1}_j$ and $E^k_j$ under two consecutive iterations are  theoretical values and cannot be known exactly in general. In practice, they are generally estimated with the gradient mean of random mini-batch samples.


\subsection{Variance of  $\bar{G}_{mst}$}
In this section, we firstly give a lemma, which we use as a basis for the discussion of the variance of $\bar{G}_{mst}$.
\begin{MyLem}\label{Lem:SPVar}
Let $E_j^{k-1}$ and $E_j^k$ be the gradient mean of the $j^{th}$ sub-population of the networks $W_{k-1}$ and $W_k$,
$V_j^{k-1}=V(G_j^{k-1})$ and $V_j^k=V(G_j^{k})$ be the gradient variance of the $j^{th}$ sub-poppulation of the networks $W_{k-1}$ and $W_k$ respectively.
Then, $\bar{G}_{mst}^k$ has a stationary point variance called $V_{sp}(\bar{G}_{mst}^k)$ as follow
\begin{equation}\label{eq:Vmin}
V_{sp}(\bar{G}_{mst}^k)=\sum\limits_j^C[\textit{w}_j^2\frac{(E_j^k)^2V_j^{k-1}\cdot V_j^{k}}{(E_j^k)^2V_j^{k-1}+(E_j^{k-1})^2V_j^{k}}]
\end{equation}
if we let
\begin{equation}\label{eq:valuepq}
p_j^k=\frac{E_j^kE_j^{k-1}V_j^k}{(E_j^k)^2V_j^{k-1}+(E_j^{k-1})^2V_j^k},
q_j^k=\frac{(E_j^k)^2V_j^{k-1}}{(E_j^k)^2V_j^{k-1}+(E_j^{k-1})^2V_j^{k}}
\end{equation}
in (\ref{eq:Gvector})
\end{MyLem}

The subscript of the left-hand term of the above formulae (\ref{eq:Vmin}) comes from the first letter of Stationary Point, not \ $min$\ as expected. That is because the function of (\ref{eq:var2}) in appendix \ref{appendix:valuepq} has only stationary point and no minimum point.

Likewise, in Lemma \ref{Lem:SPVar},the exact values of gradient expectations $E^{k-1},E^{k}$ and gradient variances $V^{k-1},V^{k}$ for the previous and subsequent iterations are generally not known. In practice, these variables are estimated by the gradient mean and variance of a random mini-batch samples.

Since the parameter \ $W_{k-1}$\  becomes another different parameter \ $W_{k}$\  after iteration, generally \ $E^{k-1}\neq E^k,V^{k-1}\neq V^k$\ . We call this phenomenon the fluctuation effect of gradient mean and variance. Many existing algorithms, such as the momentum method\cite{Qian1999}, incremental average gradient method\cite{Blatt2007}, Adam\cite{Kingma2015}, etc., do not fully consider the influence of this fluctuation effect, so the gradient direction of these algorithms are usually biased estimators.

According to (\ref{eq:valuepq}), $q_j^k < 1$ since $V_j^k\neq 0$.
However, $p_j^k$ can be great than $1$ if no further restriction.
Theorem \ref{The:p-less-one} in the next subsection gives a sufficient condition to ensure $p_j^k<1$.

\subsection{Design effect of statistics $\bar{G}_{mst}$}
The aforementioned general results are not easy to see the effect of the new statistic \ $\bar{G}_{mst}^k$\  on variance reduction. In fact, the variance of the layer (category) samples is left in each component of the vector of memory \ $G$\  by \ $\bar{G}_{mst}^k$\  in different proportions \ $p_j^k(j\in\{1,\cdots,C\},k\in\{1,2,\cdots,\})$\ , and is rapidly attenuated as the iteration proceeds. Therefore, the variance of the statistic \ $\bar{G}_{mst}^k$\ (memory type) is smaller than that of the traditional stratified sampling statistic \ $\bar{G}_{st}^k$\  (memoryless type), and the variance remaining in the memory part will be rapidly attenuated as the iteration progresses.

Since any $w_i,w_j$ in the vector $W=[w_1,w_2,\cdots,w_{\wp}]^T$ represent different connection weights in the network, the variance matrices $V_{sp}(\bar{G}_{mst}),V(\bar{G}_{st})$ of $\bar{G}_{mst}(W),\bar{G}_{st}(W)$ will be diagonal square matrices of order $\wp$.

In order to be able to compare $V_{sp}(\bar{G}_{mst}),V(\bar{G}_{st})$, the definition of the comparison of the same type of diagonal square matrix is given below.

\begin{MyDef}\label{Def:matrixleq}
Given diagonal square matrix of order $\wp$
\begin{equation}
\barwedge_1=\begin{small}\left(\begin{array}{cccccc}\sigma_1^2&&&&&\\&\sigma_2^2&&&&\\&&\bullet&&&\\&&&\bullet&&\\&&&&\bullet&\\&&&&&\sigma_{\wp}^2\end{array}\right)\end{small},
\barwedge_2=\begin{small}\left(\begin{array}{cccccc}\Sigma_1^2&&&&&\\&\Sigma_2^2&&&&\\&&\bullet&&&\\&&&\bullet&&\\&&&&\bullet&\\&&&&&\Sigma_{\wp}^2\end{array}\right)\end{small}\nonumber
\end{equation}
,if for any\ $i(i=1,2,\cdots,\wp)$\ , $\sigma_i^2\leq \Sigma_i^2$ ,then $\barwedge_1\leq \barwedge_2$
\end{MyDef}

\begin{MyDef}\label{Def:matrixle}
$\barwedge_1,\barwedge_2$\ are defined as (\ref{Def:matrixleq}) ,If there is at least one\ $i(i=1,2,\cdots,\wp)$\ ,such that\ $\sigma_i^2< \Sigma_i^2$\ ,the rest\ $j(j\neq i,j=1,2,\cdots,\wp)\sigma_i^2\leq \Sigma_i^2$\ ,then\ $\barwedge_1< \barwedge_2$\

\end{MyDef}

With the above definition of the comparison of diagonal square matrices, we can get the following Corollary \ref{cor:desigeneffect}.

\begin{MyCor}\label{cor:desigeneffect}
\ $V(\cdot)$\ represents the variance of the variable in parentheses. The statistic \ $\bar{G}_{mst}^k$\ has the following properties.
\begin{enumerate}[(1)]
\item $V_{sp}(\bar{G}_{mst}^k)<V(\bar{G}_{st}^k)$
\item $V_{sp}(\bar{G}_{mst}^{k+t})\leq p^{2t}V(\bar{G}_{mst}^{k})+\sum\limits_{i=1}^{t}p^{2(t-i)}q^2V(\bar{G}_{st}^{k+i}),0<q<1$
\end{enumerate}
\end{MyCor}

Property (1) in Corollary \ref{cor:desigeneffect} shows that the memory statistic $\bar{G}_{mst}^k$ has a smaller design effect (smaller variance) than $\bar{G}_{st}^k$.

Property (2) in Corollary \ref{cor:desigeneffect} shows that this property holds when $k$ takes any value, regardless of the starting point. Therefore, for the sake of brevity, the inequality of property (2) can be reduced to a simpler form

\begin{equation}\label{inequality:desigeneffect}
V_{sp}(\bar{G}_{mst}^{k})\leq p^{2k}V(\bar{G}_{mst})+\sum\limits_{i=1}^{k}p^{2(k-i)}q^2V(\bar{G}_{st}^{i}),0<q<1.
\end{equation}

The expansion or contraction of $V_{sp}(\bar{G}_{mst}^{k})$ mainly depends on the value of $p$. Next, we first introduce a lemma, which is used to obtain a sufficient condition to ensure that $V_{sp}(\bar{G}_{mst}^{k})$ decays rapidly with the number of iteration steps(Theorem \ref{The:p-less-one})

\begin{MyLem}\label{Lem:linearslowthanexp}
If $\eta\in (0,1)$\ , then there exists $\gamma\in (0,1)$ such that
\begin{equation}\label{inequality:linearslowthanexp}
(1+\frac{1}{1-\eta})\eta^{2k}\leq \gamma^{2(k-k_0)}
\end{equation}
holds for a sufficiently large positive integer $k,k_0,k>k_0$
\end{MyLem}

The closer $\eta$ in Lemma \ref{Lem:linearslowthanexp} is to 1, the larger the k is required to achieve the geometric decay rate on the right-hand side of (\ref{inequality:linearslowthanexp}).

\begin{MyThe}\label{The:p-less-one}
$E_j^k,E_j^{k-1},V_j^k,V_j^{k-1}$\ are defined as Lemma \ref{Lem:SPVar}. If $E_j^k=E_j^{k-1}$,then (\ref{eq:valuepq}) can be simplified as
\begin{equation}\label{The:valuepq}
p_j^k=\frac{V_j^k}{V_j^{k-1}+V_j^k},q_j^k=\frac{V_j^{k-1}}{V_j^{k-1}+V_j^{k}}.
\end{equation}
In this case,there exist $\gamma\in(0,1)$ and a sufficiently large $k_0$ such that
\begin{equation}\label{The:Vardecay}
V_{sp}(\bar{G}_{mst}^{k})\leq\gamma^{2(k-k_0)}M\cdot I ,
\end{equation}
where M is a positive constant,I is an identity matrix of order $\wp$.
\end{MyThe}

Theorem \ref{The:p-less-one}\ shows that, as long as we try to ensure that $E_j^k=E_j^{k-1}$, that is, the gradients of two adjacent iterations are expected to be equal, after a sufficiently large $k_0$ iterations, the variance of $\bar{G}_{mst}^{k}$ can decay at the geometric rate on the right side of (\ref{The:Vardecay}).

Theorem \ref{The:p-less-one} can lead to the following theorem \ref{The:VofbarGmst}.
\begin{MyThe}\label{The:VofbarGmst}
Let $Var(\bar{G}_{mst}^k):=E(||\bar{G}_{mst}^k||_2^2)-||E(\bar{G}_{mst}^k)||_2^2,\Phi=\wp M$\ ,then
\begin{equation}\label{ieq:VofbarGmst}
\begin{array}{rl}
Var(\bar{G}_{mst}^k)\leq& \gamma^{2(k-k_0)}\wp M\\
=&\gamma^{2(k-k_0)}\Phi .
\end{array}
\end{equation}
\end{MyThe}

$V_{sp}(\bar{G}_{mst}^k)$ can decay at the geometric rate of (\ref{ieq:VofbarGmst}), mainly because of the effect of the parameter $p_j^k$ (the proportion of the historical gradient being reused). This can be seen from the proof process of Theorem \ref{The:p-less-one} in Appendix \ref{appendix:plessone}. Therefore, the variance reduction strategy of $V_{sp}(\bar{G}_{mst}^k)$  is named as p-based variance reduction strategy, where the letter p of p-based refer to the maximal proportion of the historical gradient being reused,that is $p=\max p_j^k$.

Based on the result that $V_{sp}(\bar{G}_{mst}^k)$ decays at a geometric rate, we can design the corresponding MSTGD algorithm

\section{MSTGD algorithm based on $\bar{G}_{mst}$}
The update direction of \ MSTGD\  is determined by calculating the mean value of $G$ by component, which means it needs to maintain a $C$-dimension vector $G$ during iterations. At each iteration \ MSTGD\  calculates a mini-batch gradient mean $G_j$ of samples of the $j^{th}$ class, and then the $j^{th}$ component in $G$ is updated by the new $G_j$ for each class $j^{th}\in \{1,\cdots C \}$.

The pseudo code of MSTGD is described in algorithm \ref{alg:barGmstzeromean}, which is designed according to Theorem \ref{The:p-less-one} to ensure convergence.
\begin{algorithm}
\caption{\underline{M}emory-type \underline{S}tochastic s\underline{T}ratified \underline{G}radient \underline{D}escent(MSTGD) for minimizing $J(W)=\frac{1}{N}\sum_{i=1}^{N}J_i$ with step size h}
\label{alg:barGmstzeromean}
\begin{algorithmic}[1]
\STATE Parameters:Step size $h$, Batch size $B$ ,Training data size $N$, the total number of class $C$, the class weight $\textit{w}_j$,Iteration bound  $max\_iter$
\STATE Inputs:training data $\small{(x^{(1)},y^{(1)}),(x^{(2)},y^{(2)}),\cdots,(x^{(N)},y^{(N)})}$
\STATE set $G_j=0$ for $j\in\{1,2,\cdots,C\}$,k=0
\WHILE{ k$\le$max\_iter }
\FOR{$j=1,2,\cdots C $ }
\STATE calculate sample mean $E_j$ and  variance $V_j$ of $\lceil\frac{B}{C}\rceil$ samples from $j^{th}$ class
\STATE calculate $p_j^k,q_j^k$ according to equation \ref{The:valuepq}
\STATE update $j^{th}$ component of G,$G_j\leftarrow p_j\times G_j+q_j\times (g(W_k,\xi_j^k)-E_j)$, where $\xi_j^k$ is a random sample from $j^{th}$ class
\ENDFOR
\STATE $W_{k+1}=W_k-h\sum\limits_{j=1}^{C}\textit{w}_j\cdot(G_j+E_j)$
\STATE k+=1
\ENDWHILE
\end{algorithmic}
\end{algorithm}

In order to satisfy the convergence condition of \ $E_j^k=E_j^{k-1}$\ in Theorem \ref{The:p-less-one},MSTGD does not directly store the stochastic gradient $g(W_k,\xi_j^k)$  in the auxiliary variable G as in (\ref{eq:Gvector}),but first perform the mean-zeroing operation $g(W_k,\xi_j^k)-E_j$ ,and then store it in the corresponding component of G(Line 8 in Algorithm p\ref{alg:barGmstzeromean}). Our experiments show that such a mean-zeroing operation can ensure the convergence of the algorithm.

\section{Linear convergence rate of MSTGD}
This section discusses the convergence rate of MSTGD. Section \ref{sec:4.1} firstly give some basic assumptions. The theorem in Section \ref{sec:4.2} shows how the variance of the gradient in (\ref{eq:sgd-likeupdate}) specifically affects the convergence rate of the algorithm, also reveals that,if without reusing the historical gradients, due to the existence of gradient variance, the algorithms based on iteration (\ref{eq:sgd-likeupdate}) can only reach the sub-linear convergence rate. Section \ref{sec:4.3} is the linear convergence theorem of MSTGD. This theorem and its related proofs reveal the mainly reason of linear convergence of MSTGD is that the reused historical gradient greatly reduces the variance,which is called p-based variance reduction.
\subsection{Background and assumptions}\label{sec:4.1}
To build the general convergent result we need the following assumptions.
\begin{itemize}
\item $A_1)$ The Cost function $J(W)$ is continuously differentiable and  first order Lipschitz continuous with Lipschitz constant $L>0$, i.e.,\begin{equation}\label{eq:11}
\parallel \nabla J(W)-\nabla J(W')\parallel \leq L\parallel W-W'\parallel.
\end{equation}
\item $A_2)$ Cost function $J(W)$ is strongly convex, i.e.,\begin{equation}\label{eq:12}
J(W')\geq J(W)+\nabla J(W)(W'-W)+\frac{1}{2}c\parallel W'-W\parallel^2.
\end{equation}
This assumption leads to a useful fact(proved in Appendix \ref{proof:strongconvex})
\begin{equation}\label{ieq:strongconvex}
2c(J(W)-J_*)\leq||\nabla J(W)||^2_2.
\end{equation}
\item $A_3)$ The objective function J and stochastic gradient\ $g(w_k,\xi_j^k)$\ satisfy the following conditions:
\begin{enumerate}
\item The sequence of iterates\ $\{W_k\}$\ is contained in an open set over which J is bounded below by a scalar\ $J_{inf}$
\item There exist scalars\ $\mu_G\geq\mu>0$\ such that, for all\ $k\in\{1,2,\cdots\} $\ ,
\begin{equation}
\nabla J(W_k)^TE[\bar{G}_{mst}^k]\geq \mu||\nabla J(W_k)||_2^2,
\end{equation} and
\begin{equation}
||E[\bar{G}_{mst}^k]||_2\leq \mu_G||\nabla J(W_k)||_2.
\end{equation}
\end{enumerate}
\end{itemize}
\subsection{General convergent result of gradient descent}\label{sec:4.2}
Before analyzing the convergence rate of MSSG, we present a general convergent result of GDM (Gradient Descent Methods) in this section, where GDM refers specifically to FGD, SGD and mini-batch SGD. From this general result, we can see how gradient variance impacts the convergence rate of an  algorithm.\\

Theorem \ref{th:cvi} formalizes the relationship between gradient variance and convergence rate. With smaller gradient variance, GDM gets closer to the optimal solution, and if gradient variance is reduced to zero, GDM can achieve linear convergence rate.
\begin{MyThe}[Convergence-Variance Inequality,CVI]\label{th:cvi}
Let\ $W_{k}$\ be a network obtained by a method of GDM,$\sigma_k^2(s_k^2)$ is the gradient variance on population(samples) at the $k^{th}$ iteration,\ $\sigma^2_{min}=\min\{\sigma^2_k\}$,$\sigma^2_{max}=\max\{\sigma^2_k\}$\ .If assumptions $A1)$ and $A2)$ hold, then under the condition of step size $h_k\!<\!\frac{2}{L}$ and the ratio of sample size $\frac{n_{k+1}}{n_k}\geq \frac{\sigma^2_{min}}{\sigma^2_{max}}\frac{h_{k+1}}{h_k}\frac{2-h_kL}{2-h_{k+1}L}$, the following inequality holds for all $k\in\{1,2,\cdots\}$:
\begin{equation}
E[J(W_{k+1})-J_*]\!<\! \Lambda_1+\rho^k(E[J(W_1)-J_*]-\Lambda_1),\nonumber
\end{equation}
where $J_*$ is the optimal value, $\Lambda_k\!=\!\frac{h_kLs_k^2}{2\mu(2-h_kL)}$,$\rho_i:=1-\mu h_i(2-Lh_i),\rho=\max\{\rho_i\}$.
\end{MyThe}

CVI theorem is a general result of FGD, SGD, and Batch. In the case of FGD, the sample size n is equal to data size (population size) N; the sampling ratio f equals $1$, so $s_k^2=0$ ,also $\Lambda_1=0$. This leads to linear convergence rate of FGD. In the case of SGD, the sample size n is equal to one, the number $\Lambda_1$ ceases to decay.p In this case, SGD cannot achieve a linear convergence rate, it can only be a slower sub-linear convergence rate. As for Batch, the sample size n is a random number between $1$ and N. Therefore, if a strategy of cleverly setting the size of n, such as a dynamic sample size n that satisfies certain conditions, the number $\Lambda_1$ can be reduced infinitely close to zero, Batch still retains the possibility of linear convergence.



This paper presents Theorem \ref{th:cvi} and its proof independently (see Appendix \ref{appendix:cvi}), but Theorem \ref{th:cvi} is not our first creation, similar works and related results can be found in earlier literature\cite{Bottou2018,nesterov2013introductory}

\subsection{Convergence results of MSTGD}\label{sec:4.3}
Theorem \ref{th:cvi} shows that for the GDM using the iteration of (\ref{eq:sgd-likeupdate}), since no historical gradient is used, even if a strategy such as dynamic step size is employed, the linear convergence rate cannot generally be achieved. Only in special cases, such as using a dynamic sample size strategy, can a linear rate of convergence be achieved.

Theorem \ref{th:linear} pioneered in this paper and its corresponding proof (see Appendix \ref{appendix:linear}) show that the MSTGD using the iteration of (\ref{eq:weightupdate})and (\ref{eq:Gvector}) can achieve linear convergence rate with a constant stepsize and constant sample size due to the efficient reuse of historical gradient.
\begin{MyThe}[Linear Convergence of MSTGD]\label{th:linear}
Suppose that Assumptions $A_1),A_2)$, and $A_3)$(with $J_{inf}=J_*$) hold, In addition, MSTGD is run with a fix stepsize $h_k=\bar{h},\forall k\in \{1,2,\cdots\}$, and satisfying
\begin{equation}
0\le \bar{h}\leq min\{\frac{\mu}{L\mu_G^2},\frac{1}{c\mu}\},
\end{equation}
then for all $k\in \{1,2,\cdots\}p$, the expected optimality gap satisfies
\begin{equation}\label{The:linarconvergence}
E[J(W_k)-J_*]\leq\Omega\lambda^{2(k-k_0)},
\end{equation}
where
\begin{equation}
\Omega:=max\{\frac{\bar{h}L\Phi}{c\mu},J(W_1)-J_*\},
\end{equation} and
\begin{equation}
\lambda:=max\{1-\frac{\bar{h}c\mu}{2},\gamma\}<1.
\end{equation}

\end{MyThe}


Theorem \ref{th:linear} shows that after a sufficiently large $k_0$ iterations, MSTGD can linearly converge to the optimal solution at the rate of the right-hand side of (\ref{The:linarconvergence}).

\section{Experimental results}\label{sec:experimence}
We make comparison among $\bar{G}_{mst}$, $\bar{G}_{st}$, $SGD$ and $Batch$,
by testing them on artificial data as well as the MNIST data set.
It turns out that $\bar{G}_{mst}$ has the higher estimation accuracy than the other methods.

The artificial data is generated in the form of a $40\times 10$ random matrix
(that is, $40$ random numbers in each round, and $10$ rounds successively, each random number is a simulation of a random gradient $g(W_k,\xi_j^k)$).
Also, various kinds of data set are generated so that they are
with increasing mean, decreasing mean, increasing variance, and decreasing variance
in successive rounds, respectively.
In each round, we divide the $40$ numbers into $4$ sub-populations
as a layered simulation.
In the experiment, we take the square of the deviation between the estimated value generated by the estimator
and the overall mean (true value) as an evaluation of the accuracy of the search direction provided by the estimator.

For fair and comparable consideration,
the number of samples for each method is set to $4$, except for $SGD$ which uses a single sample.
$\bar{G}_{mst}$ and $\bar{G}_{st}$ randomly select a sample from each of the $4$ sub-populations,
and $Batch$ randomly select $4$ samples from the whole population.

The parameters $p_j^k$ and $q_j^k$ in $\bar{G}_{mst}$ are calculated according to formulae (\ref{eq:valuepq}),
where $E_j$ and $V_j$ respectively takes the value of the mean and variance of the sub-populations in the random data set.

\subsection{Results on a uniformly random data set}
In the first experiment, we uniformly sample from each of the following $10$ intervals $$[8,12],[8,10],[6,9],[5,8],[4,7],[3,6],[3,5],[2,4],[2,3],[0,3]$$
(with decreasing mean) to form a random data set with a specification of $40 \times 10$ as the data population.

\begin{figure}[htbp]
  \centering
    \includegraphics[width=\textwidth]{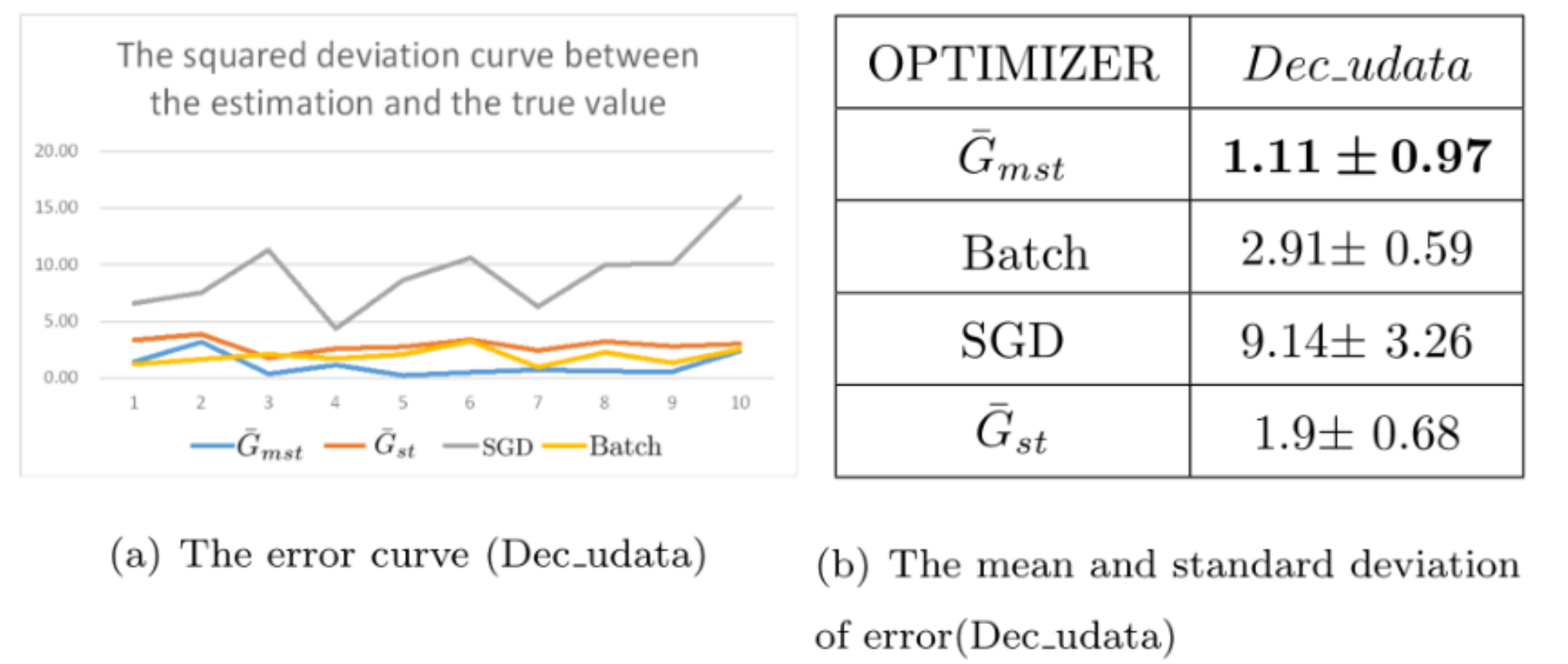}
    \caption{\small{the error curve and its description statistics(Dec$\_$udata)}}
    \label{fig:randomdec}
\end{figure}

The error in this paper is uniformly expressed by the square of the deviation between the estimator and the true value.
Figure \ref{fig:randomdec} (a) and (b) show the error curve of each estimator and the corresponding error descriptive statistics.
As can be seen in Figure \ref{fig:randomdec} (a), the blue curve corresponding to $\bar{G}_{mst}$ is located at the bottom,
under all other curves, indicating that the estimated value provided by $\bar{G}_{mst}$  statistic is the closest to the true value.
Figure \ref{fig:randomdec} (b) shows that the error of $\bar{G}_{mst}$  has the smallest mean and standard deviation,
indicating that the estimated value provided by $\bar{G}_{mst}$ is more accurate and more stable.

\begin{figure}[htbp]
  \centering
    \includegraphics[width=\textwidth]{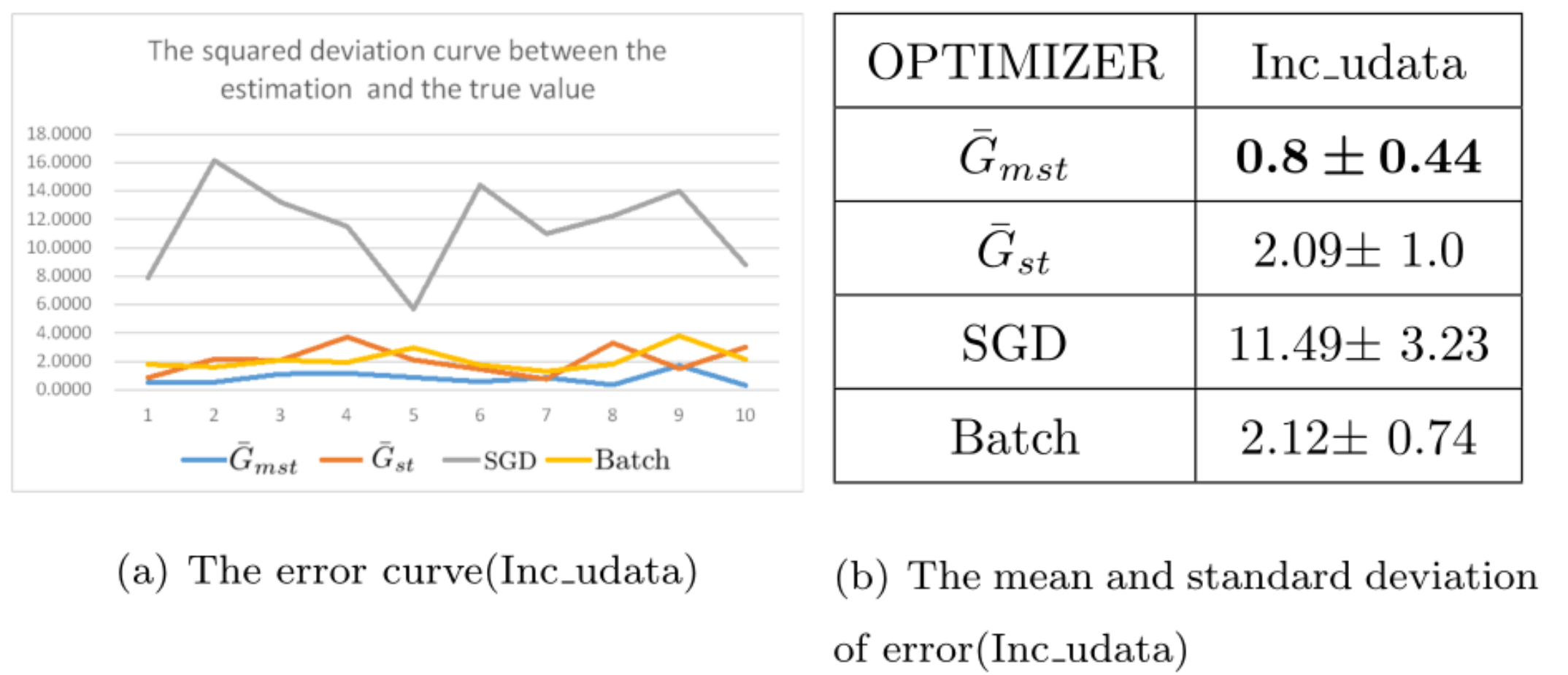}
    \caption{\small{the error curve and its description statistics(Inc$\_$udata)}}
    \label{fig:randominc}
\end{figure}

The second random data set is uniformly generated from ten intervals
$$[0,3],[2,3],[2,4],[3,5],[3,6],[4,7],[5,8],[6,9],[8,10],[8,12],$$
with increasing mean, which is also formatted into a $40\times 10$ matrix as the data population.

It can be seen from Figure \ref{fig:randominc} that on the mean increasing data set,
the search direction provided by $\bar{G}_{mst}$ is still satisfactory.
For the increasing mean data set, in Figure \ref{fig:randominc} (a),
the blue curve corresponding to $\bar{G}_{mst}$ is also located at the bottom, under of all other curves.
In Figure \ref{fig:randominc} (b), the error of $\bar{G}_{mst}$ also has the smallest mean and standard deviation,
so the estimated value provided by $\bar{G}_{mst}$ is more accurate and more stable.

\subsection{Results on normal random data set}
To further investigate the performance of the algorithms on other kind of data sets,
we generate sets of $40\times 10$ random numbers from the normal population $N(\mu,\sigma)$ as experimental data.
According to the conventional way of experimental data design,
the following five different types of data sets are generated.
\begin{enumerate}
\item Random data set where $\mu,\sigma$ are both uniformly distributed random numbers
in the interval $[1,20]$, denoted by random\_ndata
\item Random data set with decreasing mean $\mu$, denoted by MeanD\_ndata
\item Random data set with increasing mean $\mu$, denoted by MeanI\_ndata
\item Random data set with decreasing variance, denoted by VarD\_ndata
\item Random data set with increasing variance, denoted by VarI\_ndata
\end{enumerate}

\begin{figure}[htbp]
  \centering
    \includegraphics[width=\textwidth]{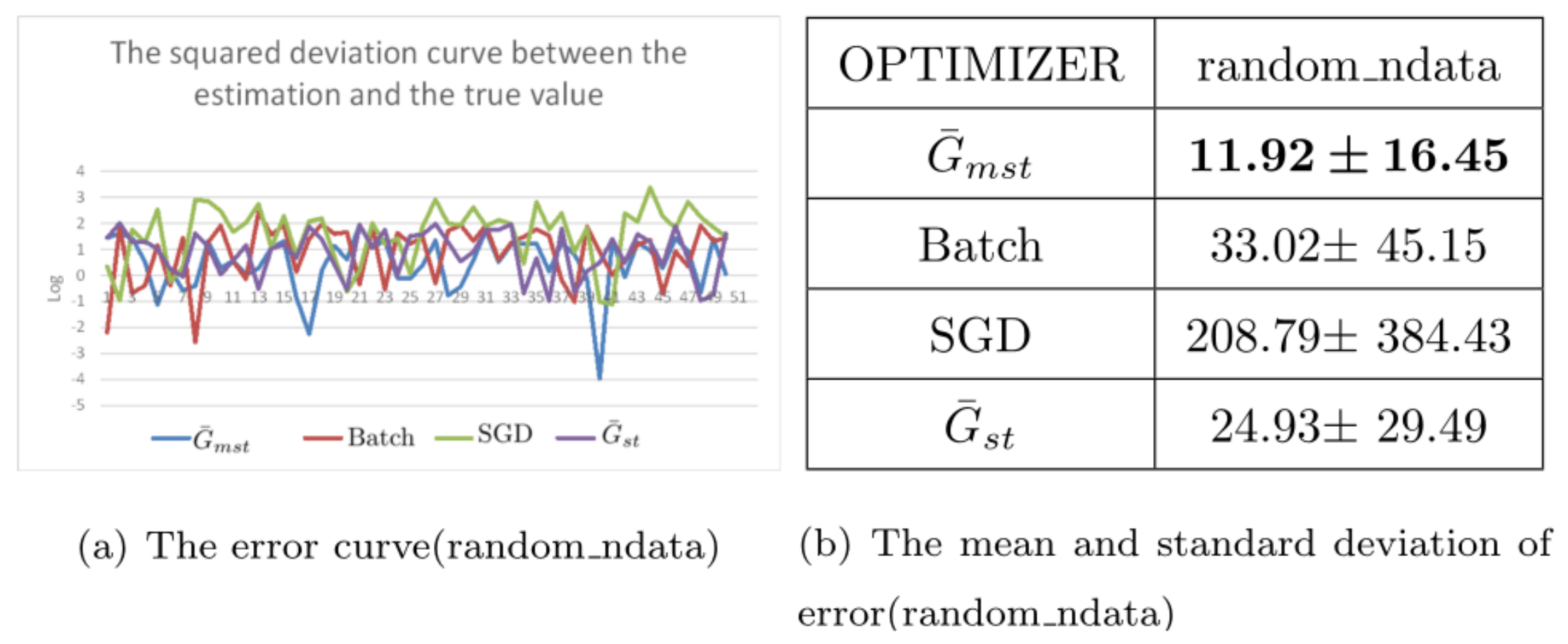}
    \caption{\small{the error curve and its description statistics(random$\_$ndata)}}
    \label{fig:randomn}
\end{figure}

\begin{figure}[htbp]
  \centering
    \includegraphics[width=\textwidth]{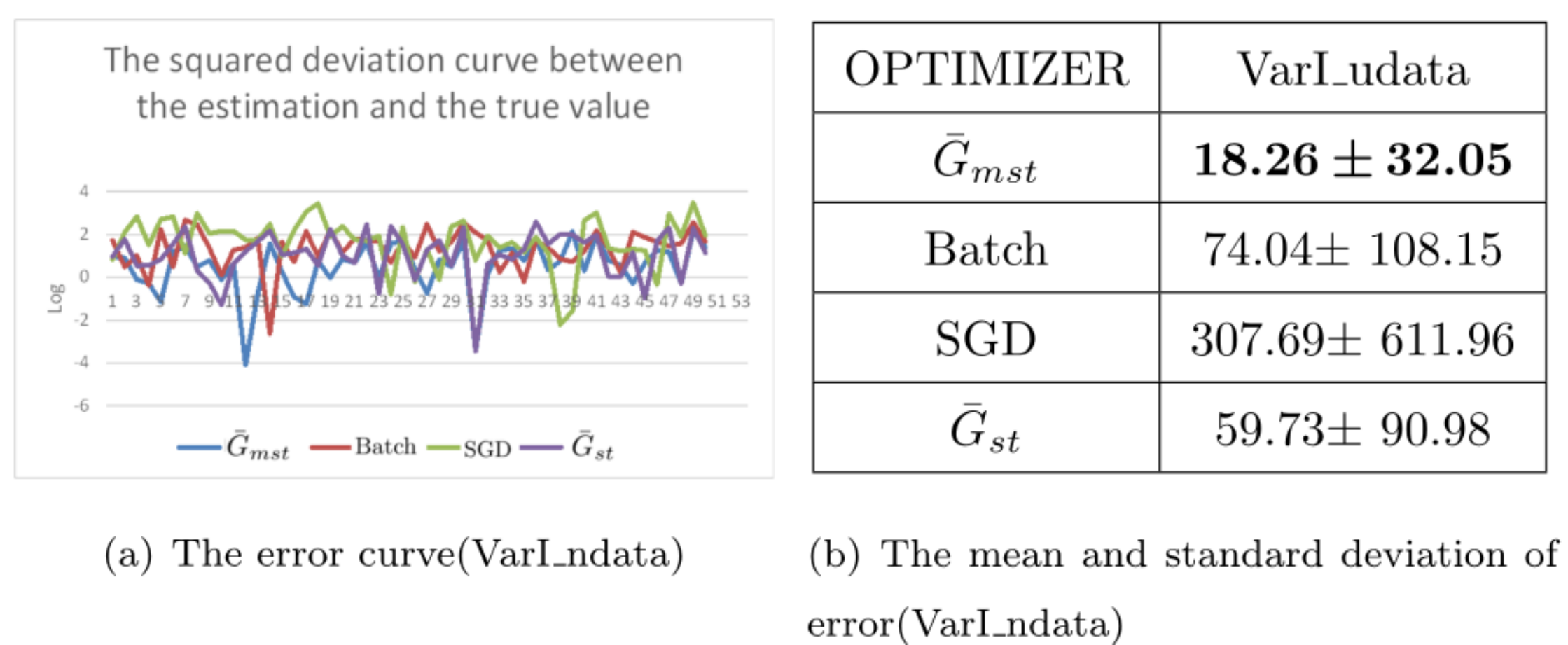}
    \caption{\small{the error curve and its description statistics(VarI$\_$udata)}}
    \label{fig:vari}
\end{figure}

The experimental results are summarized in Figure \ref{fig:randomn}$\sim$\ref{fig:vari}.
It can be seen that, in experiment on each type of data,
the gradient estimation value generated by the statistic $\bar{G}_{mst}$
always has the smallest error and smallest standard deviation.
This shows that the statistic $\bar{G}_{mst}$ can approximate the gradient mean (true value)
of population more accurately and more stably.

\begin{figure}[htbp]
  \centering
    \includegraphics[width=\textwidth]{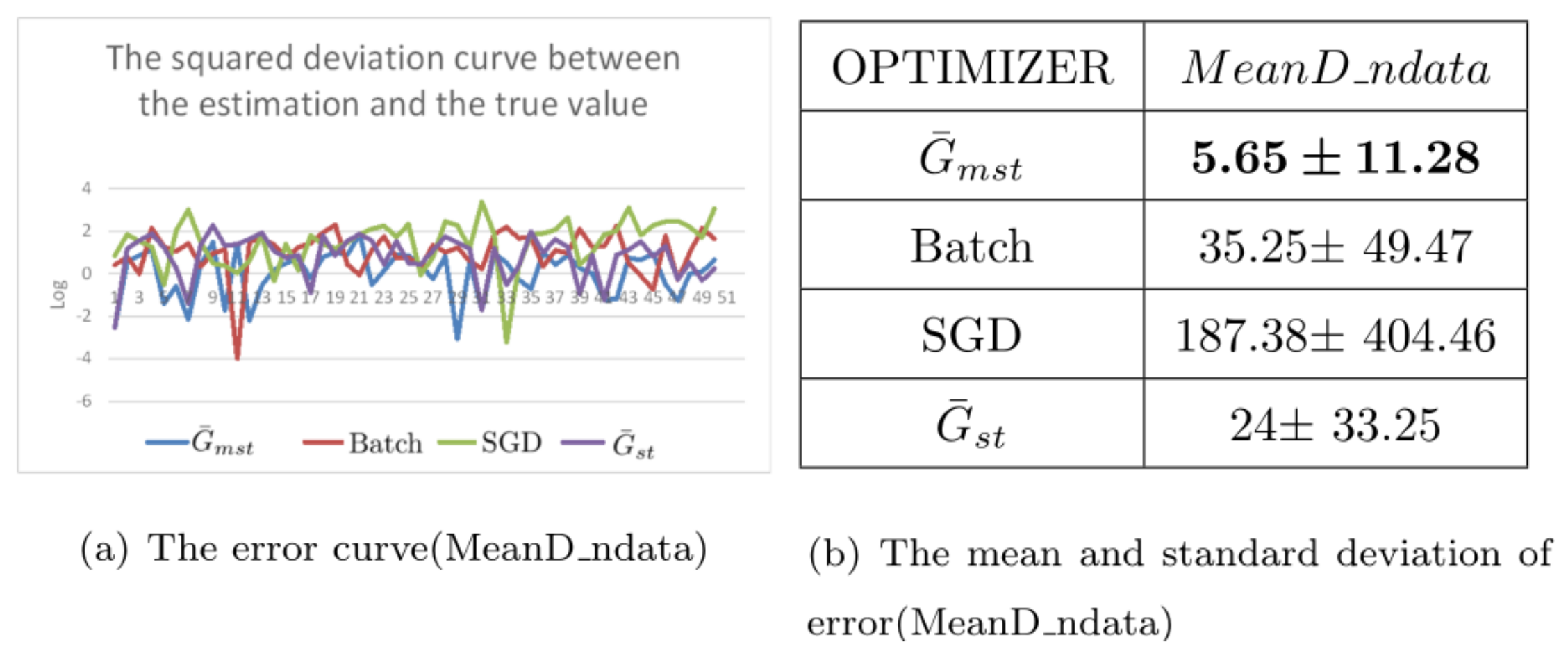}
    \caption{\small{the error curve and its description statistics(MeanD$\_$ndata)}}
    \label{fig:meand}
\end{figure}

\begin{figure}[htbp]
  \centering
    \includegraphics[width=\textwidth]{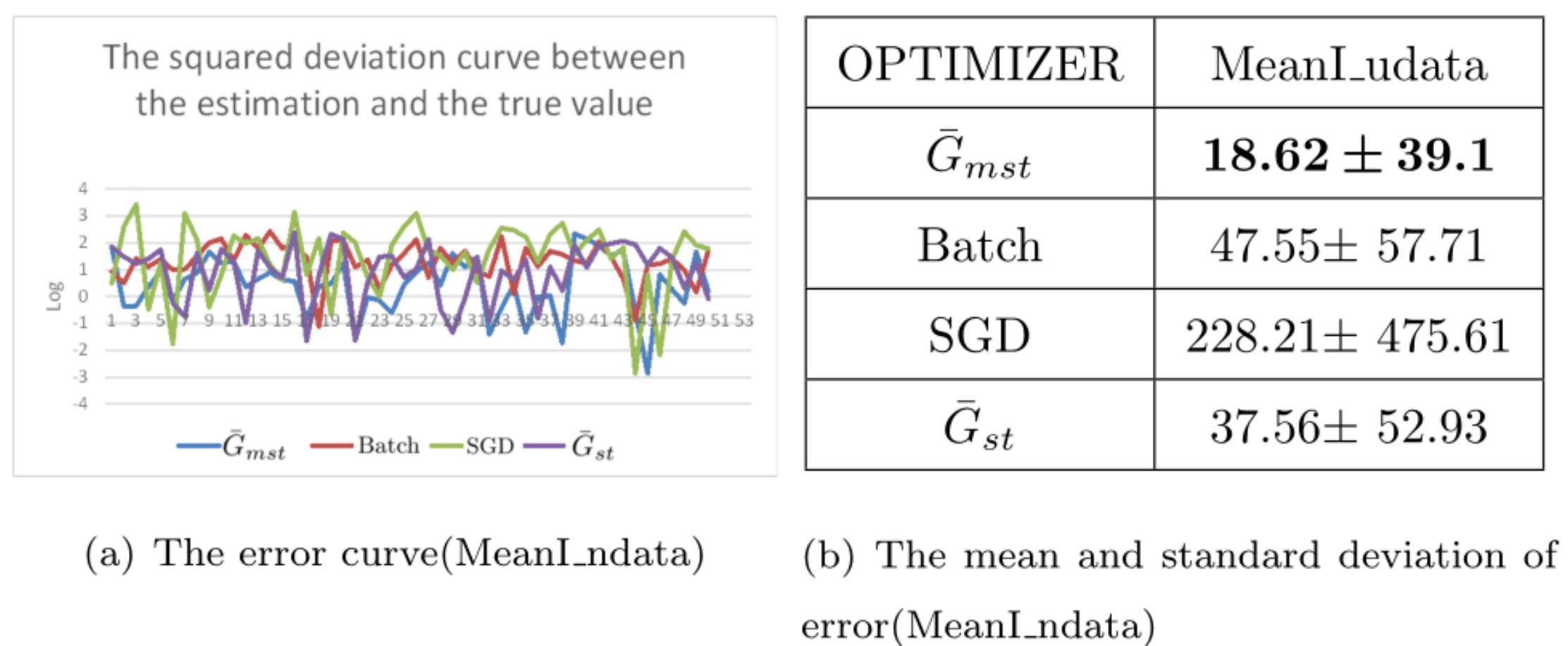}
    \caption{\small{the error curve and its description statistics(MeanI$\_$udata)}}
    \label{fig:meani}
\end{figure}

\begin{figure}[htbp]
  \centering
    \includegraphics[width=\textwidth]{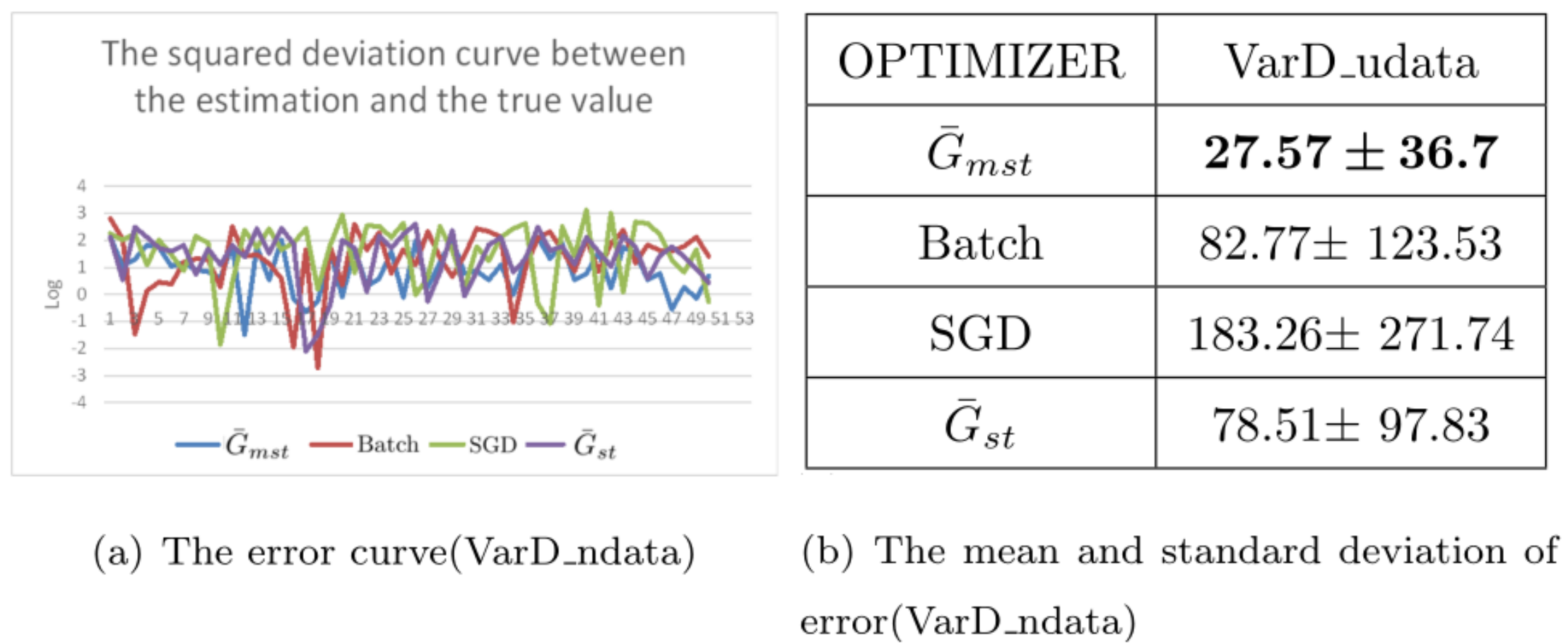}
    \caption{\small{the error curve and its description statistics(VarD$\_$udata)}}
    \label{fig:vard}
\end{figure}

\subsection{Results on the MNIST dataset}
Next we compare the performance of different algorithms on the MNIST data set.
we use the full gradient method to train a $5$-layer forward network with a structure
of $[784,500, 500,200,10]$, a training set of $60,000$ scale,
step size $h=0.2$, and weight decay coefficient $\lambda=0.001$.
After $60$ iterations, the network achieves $87.73\%$ accuracy on the $10,000$-scale test set.
We record the weight $w$ between the first neuron in the output layer and the first neuron in the penultimate layer
as the gradient information about $60,000$ training samples in 60 iterations,
forming a gradient matrix with a scale of $60,000\times 60$.
The average value of each column of the gradient matrix is the true gradient direction of the parameter update.

On the $60,000\times 60$ gradient matrix,
we calculate the expected $E_j$ and variance $V_j$ of category $j$,
then we calculate $p_j,q_j$ according to formulae (\ref{eq:valuepq}),
where $j$ runs from $0$ to $9$.
Finally, the value of the $\bar{G}_{mst}$ can be calculated using these parameters.

For the sake of comparability, except that $SGD$ uses a single sample,
the sample sizes of $Batch,\bar{G}_{mst},\bar{G}_{st}$ are all set to $10$.
Among them, $\bar{G}_{mst}$ and $\bar{G}_{st}$ randomly select a single sample from each of the $10$ categories
and Batch randomly selects $10$ samples from the $60,000$ samples.

\begin{figure}
\centering
\subfigure[$\bar{G}_{mst}$ VS. Pop.]{\includegraphics[width=0.45\textwidth]{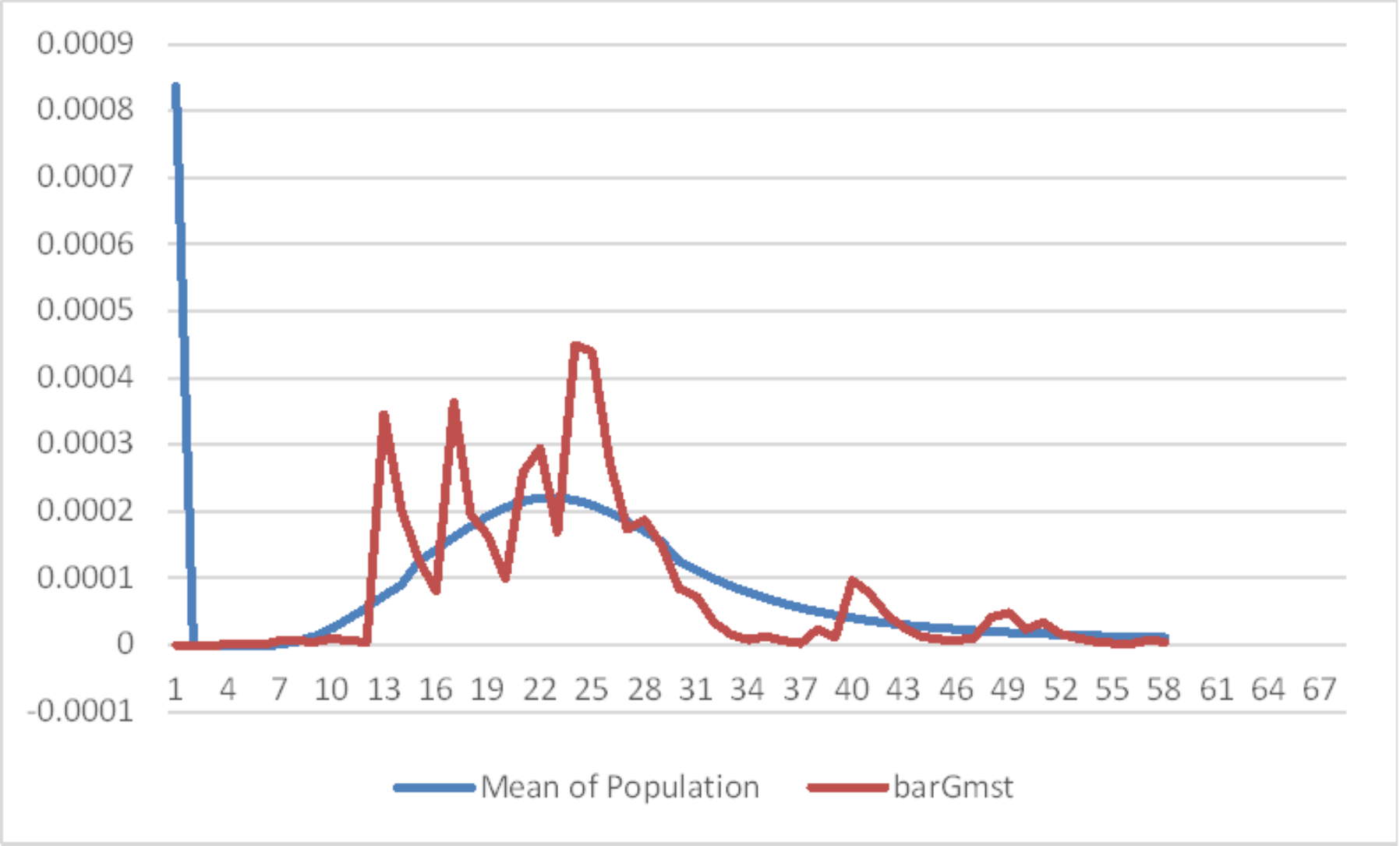}}
\subfigure[$\bar{G}_{st}$ VS. Pop.]{\includegraphics[width=0.45\textwidth]{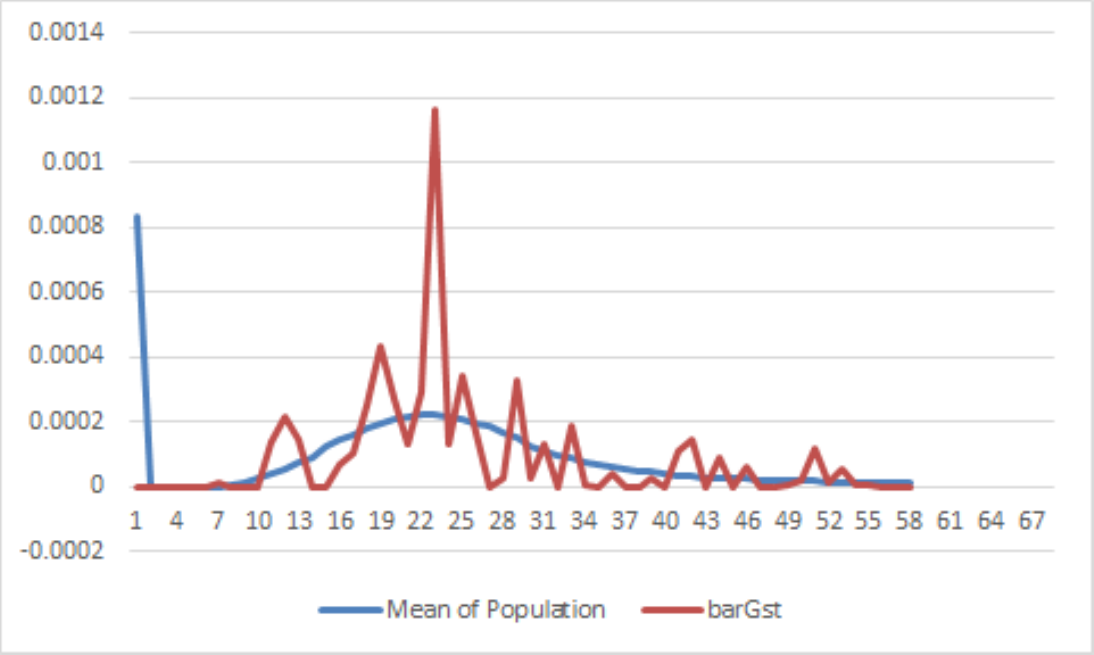}}
\\
\centering
\subfigure[Batch VS. Pop.]{\includegraphics[width=0.45\textwidth]{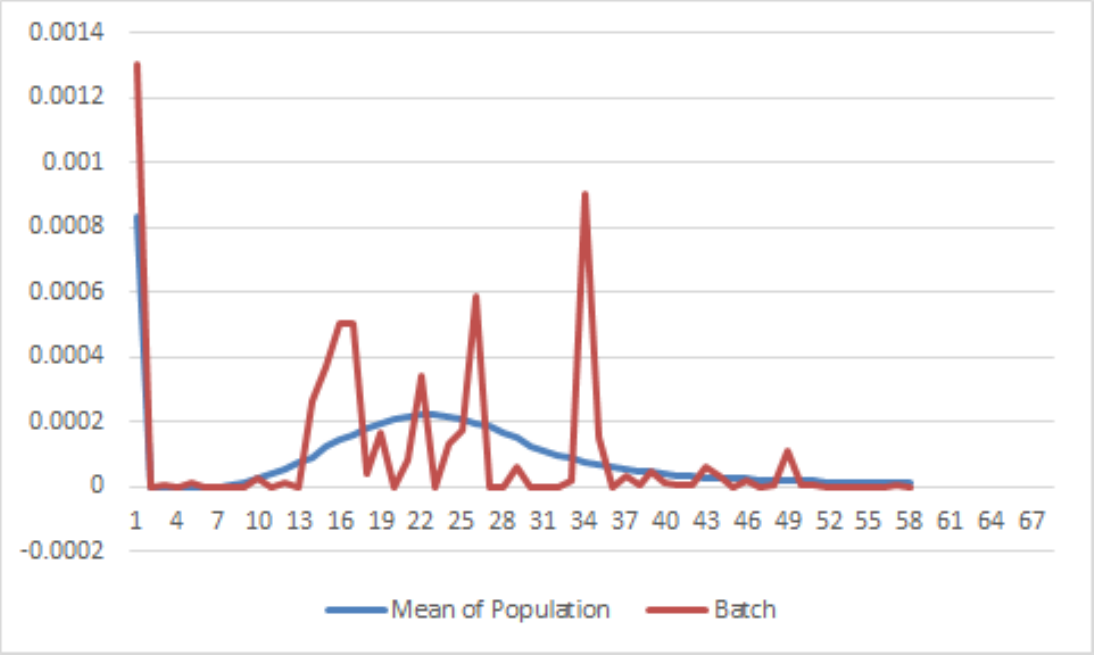}}
\subfigure[SGD VS. Pop.]{\includegraphics[width=0.45\textwidth]{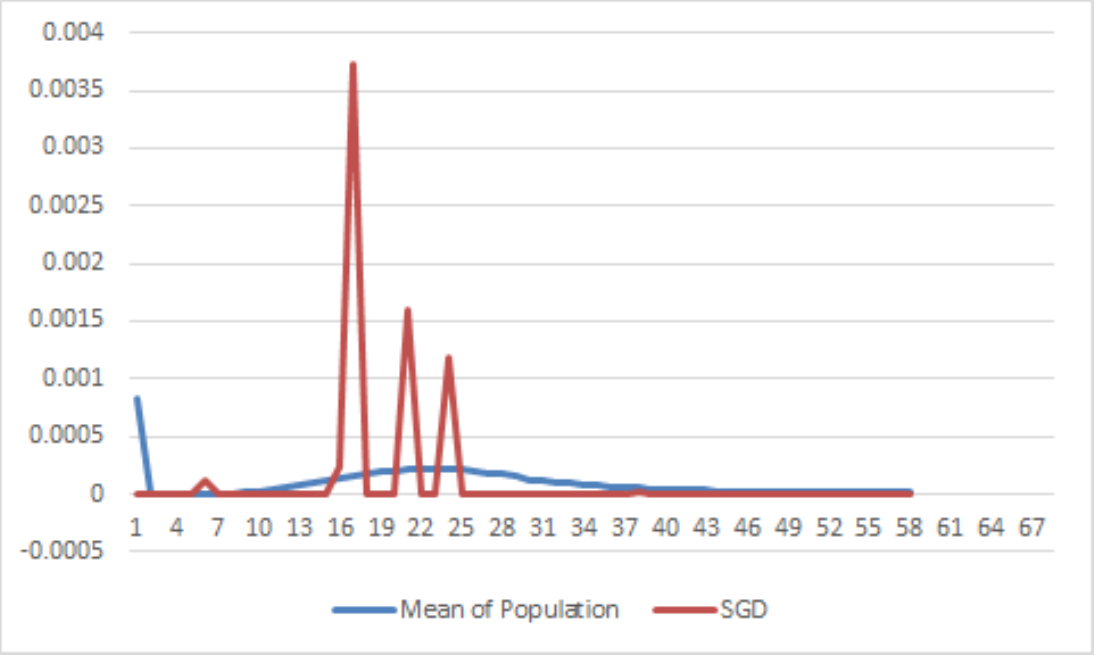}}
\caption{Tracing curves of four different estimators to true values}
\label{fig:mnist}
\end{figure}

The blue curve in Figure \ref{fig:mnist} represents the true gradient direction
(denoted by ``Pop" in Figure \ref{fig:mnist}),
which is the average gradient sequence generated by the full gradient method after $60$ iterations.
The red curve in each part stands for the curve formed by the gradient sequence generated after $60$ iterations of different algorithms.
The four graphics show the subtle differences in tracking the direction of the blue curve by the gradient curves generated by different methods.

In order to further investigate the accuracy difference of the four methods of \ $Batch,SGD,\bar{G}_{mst},\bar{G}_{st}$\ , we calculate and record the deviation square of the random gradient direction generated by the four methods and the true gradient direction. Thus, each method obtains 60 such deviation squares, and the average and standard deviation of these 60 deviation squares are used to measure the performance of the algorithm.

\begin{figure}[htbp]
\centering
\includegraphics[width=0.8\textwidth]{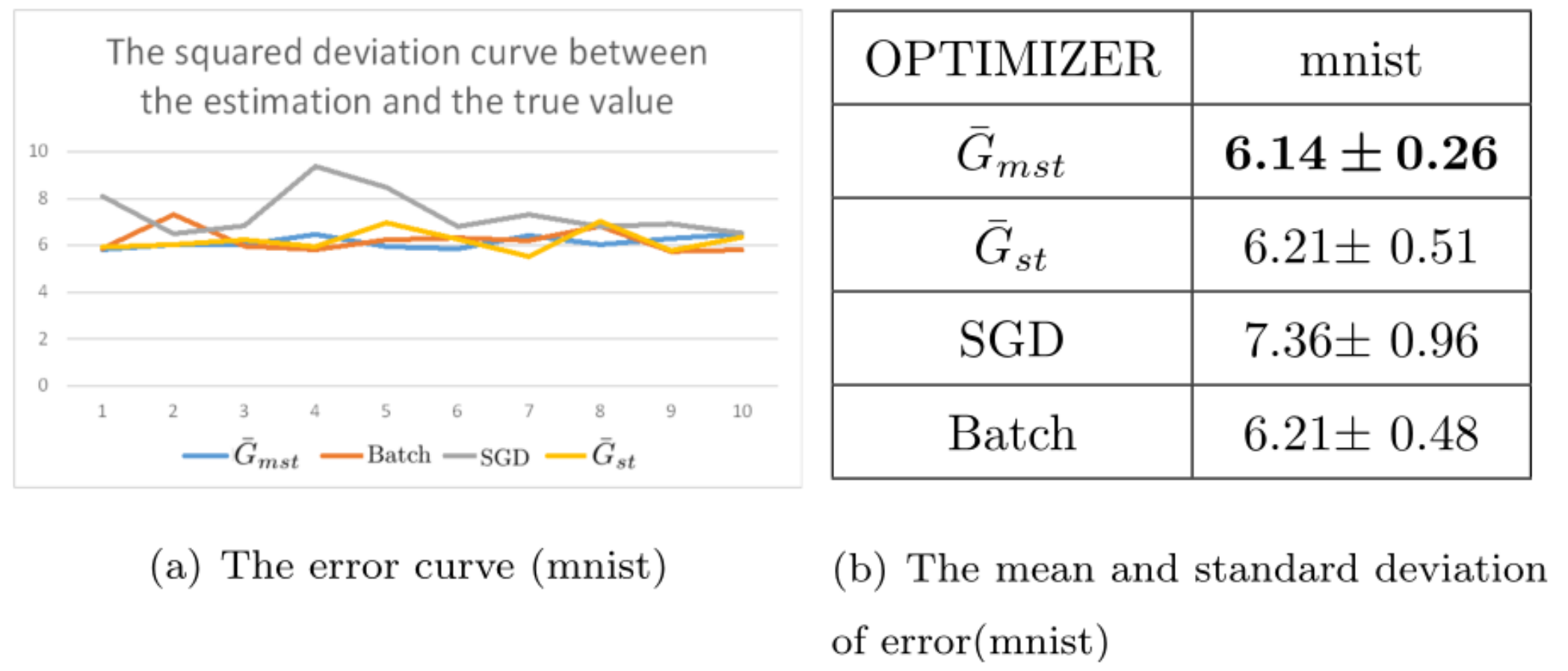}
\caption{\small{the error curve and its description statistics(mnist)}}
\label{fig:mnist2}
\end{figure}

Each algorithm is run repeatedly for 10 times, and the deviation square values generated  are recorded. It can be seen from Figure \ \ref{fig:mnist2}\  that even in the case of a small sampling ratio \ $f=\frac{10}{60000}$\ , the search direction provided by \ $\bar{G}_{mst}$\  is closer to the true value than the other methods
(the mean and standard deviation of the deviation square are the smallest).

we further compare the performance of the four algorithms Batch, SGD, MSTGD, and $\bar{G}_{st}$
in optimizing a $5$-layer forward network $[784,500, 500,200,10]$. The data used is still the $60,000$-scale MNIST training set and the $10,000$-scale MNIST test set. But to be fair, we first implement a grid search procedure for the optimal hyperparameters in the range of h=[0.01,1,0.001], $\lambda$=[0.001,0.0001],then examine the training and test accuracy differences of the four algorithms  after 1, 2, 3, 4, 5, 6, 7, 8, 9, and 10 thousand iterations under their respective optimal hyperparameters.

Table\ \ref{tab:traintestaccu}\  shows the training and testing accuracy achieved by the four algorithms Batch,SGD,MSTGD,$\bar{G}_{st}$\  under their respective optimal hyperparameters. As can be seen from Table \ \ref{tab:traintestaccu}\ , except for 7 and 9 thousand iterations, the training and the test accuracy of \ $\bar{G}_{mst}$\  are the best, outperform than the other four algorithms.
\begin{table}[htbp]
  \centering
  \caption{the training and testing accuracy achieved by Batch,SGD,MSTGD,$\bar{G}_{st}$\  }
  \scalebox{0.7}{
    \begin{tabular}{|c|c|c|c|c|c|c|c|c|}
    \hline
    Iterations & \multicolumn{2}{c|}{SGD(\%)} & \multicolumn{2}{c|}{MSTGD(\%)} & \multicolumn{2}{c|}{Batch(\%)} & \multicolumn{2}{c|}{$\bar{G}_{st}$(\%)} \\
    \cline{2-9}
        ($10^3$)  & test accu & train accu & test accu & train accu & test accu & train accu & test accu & train accu \\
        \hline
    1(SGD:$\times$20) & 91.48 & 91.39 & \textbf{94.46} & \textbf{94.56} & 93.31 & 93.38 & 94.12 & 94.28 \\
    2(SGD:$\times$20) & 92.87 & 92.94 & \textbf{96.05} & \textbf{96.37} & 95.52 & 95.88 & 95.92 & 96.18 \\
    3(SGD:$\times$20) & 93.83 & 94.08 & \textbf{96.53} & \textbf{96.93} & 96.35 & 96.86 & 96.17 & 96.42 \\
    4(SGD:$\times$20) & 94.57 & 94.76 & \textbf{96.74} & \textbf{97.39} & 96.61 & 97.25 & 96.44 & 96.78 \\
    5(SGD:$\times$20) & 94.7  & 94.79 & \textbf{97.17} & \textbf{97.98} & 97.12 & 97.81 & 96.65 & 97.18 \\
    6(SGD:$\times$20) & 95.26 & 95.8  & \textbf{97.3}  & \textbf{98.11} & 97.16 & 97.91 & 96.94 & 97.39 \\
    7(SGD:$\times$20) & 95.05 & 95.7  & 97.13 & \textbf{98.19} & \textbf{97.23} & 97.89 & 96.64 & 97.22 \\
    8(SGD:$\times$20) & 95.55 & 95.82 & \textbf{97.65} & \textbf{98.4}  & 97.43 & 98.33 & 96.62 & 97.09 \\
    9(SGD:$\times$20) & 95.66 & 96.24 & 97.41 & 98.41 & \textbf{97.72} & \textbf{98.49} & 96.92 & 97.42 \\
    10(SGD:$\times$20) & 95.44 & 95.9  & \textbf{97.63} & \textbf{98.81} & 97.4  & 98.45 & 97.31 & 97.81 \\
    \hline
    \end{tabular}%
    }
  \label{tab:traintestaccu}%
\end{table}%
\subsection{Results on Convergence rate}
\begin{figure}[htbp]
\centering
\includegraphics[width=0.8\textwidth]{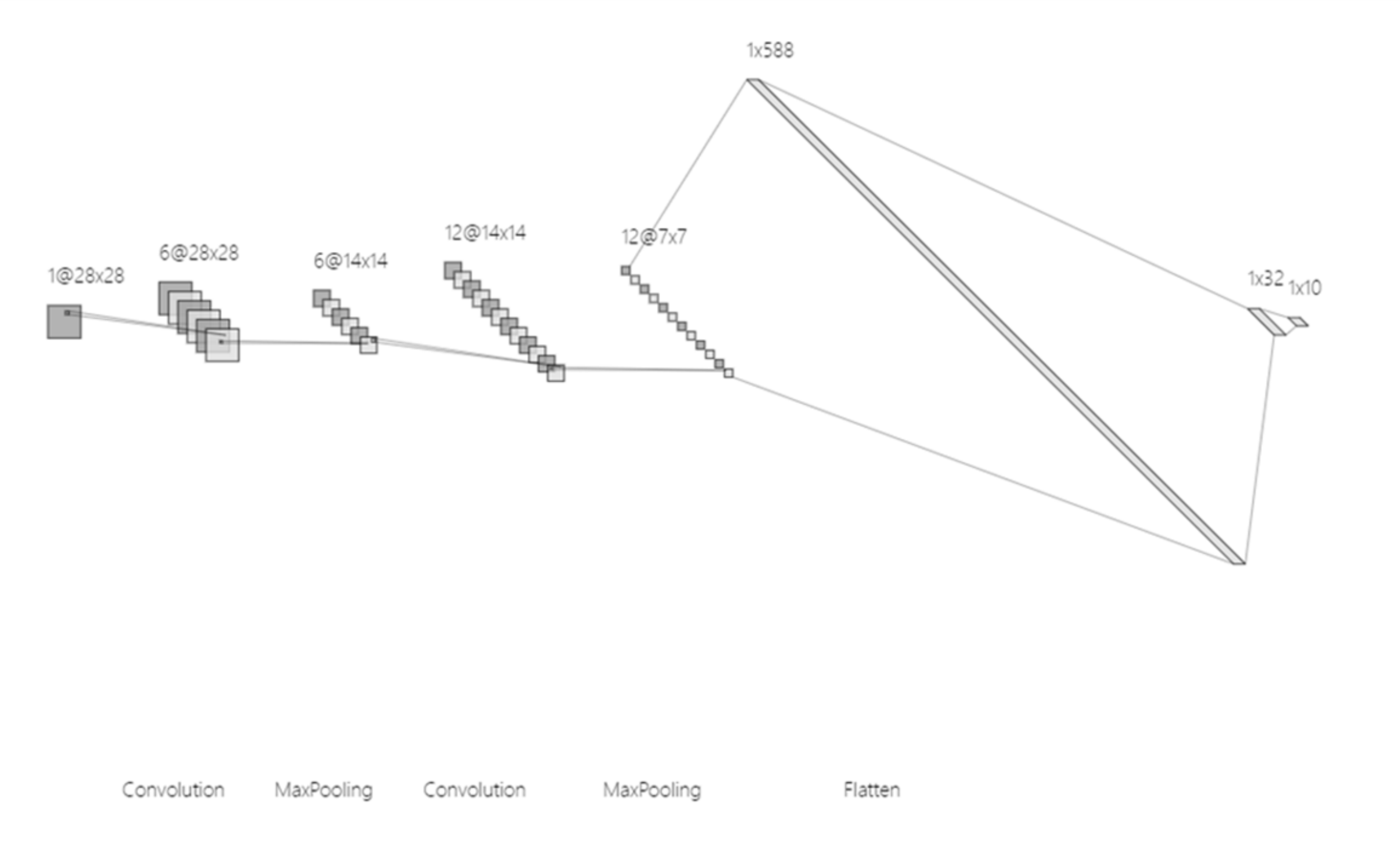}
\caption{\small{The convolutional neural network structure used in this paper to test the convergence performance}}
\label{fig:cnn}
\end{figure}

An author of this paper independently tested the convergence performance of MSTGD on BP feedforward networks and convolutional networks using a four-layer feedforward network of [784, 64, 32, 10], and the form of Fig. \ref{fig:cnn} Convolutional Neural Networks.


To be fair, whether it is a BP forward network or a CNN convolutional network, during training, except for SGD, the Batchsize of the other three algorithms is 20. Also to be fair, except for SGD, the other three algorithms are the test accuracy recorded after every 1000 iterations, while SGD is the test accuracy after every 20*1000=20,000 iterations. For each algorithm, the network was trained 14 times with the MNIST dataset, and the average test accuracy of the 14 times was recorded.

\begin{figure}[htbp]
\centering
\includegraphics[width=0.8\textwidth]{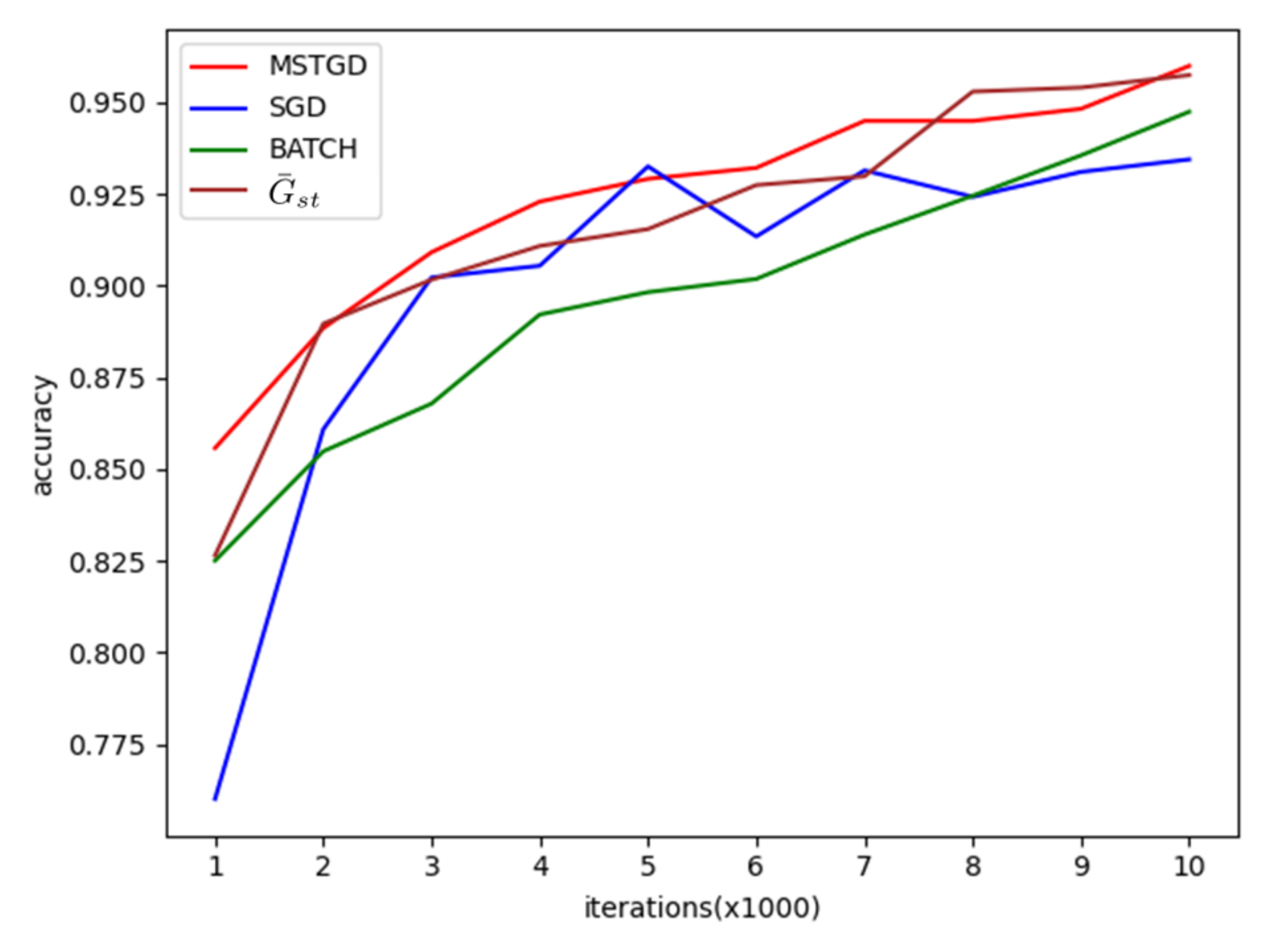}
\caption{\small{(BP on MNIST)Convergence rate of MSTGD,SGD,Batch,$\bar{G}_{st}$}}
\label{fig:Linearonbp}
\end{figure}

Figure \ref{fig:Linearonbp} is the change curve of the average test accuracy of the four algorithms MSTGD, SGD, Batch, $\bar{G}_{st}$ when training the BP forward network for 14 times. As can be seen from Figure \ref{fig:Linearonbp}, compared with other algorithms, MSTGD can obtain the best test accuracy under the same number of iterations. This shows that MSTGD has a faster convergence rate when training the forward network with MNIST data.

\begin{figure}[htbp]
\centering
\includegraphics[width=0.8\textwidth]{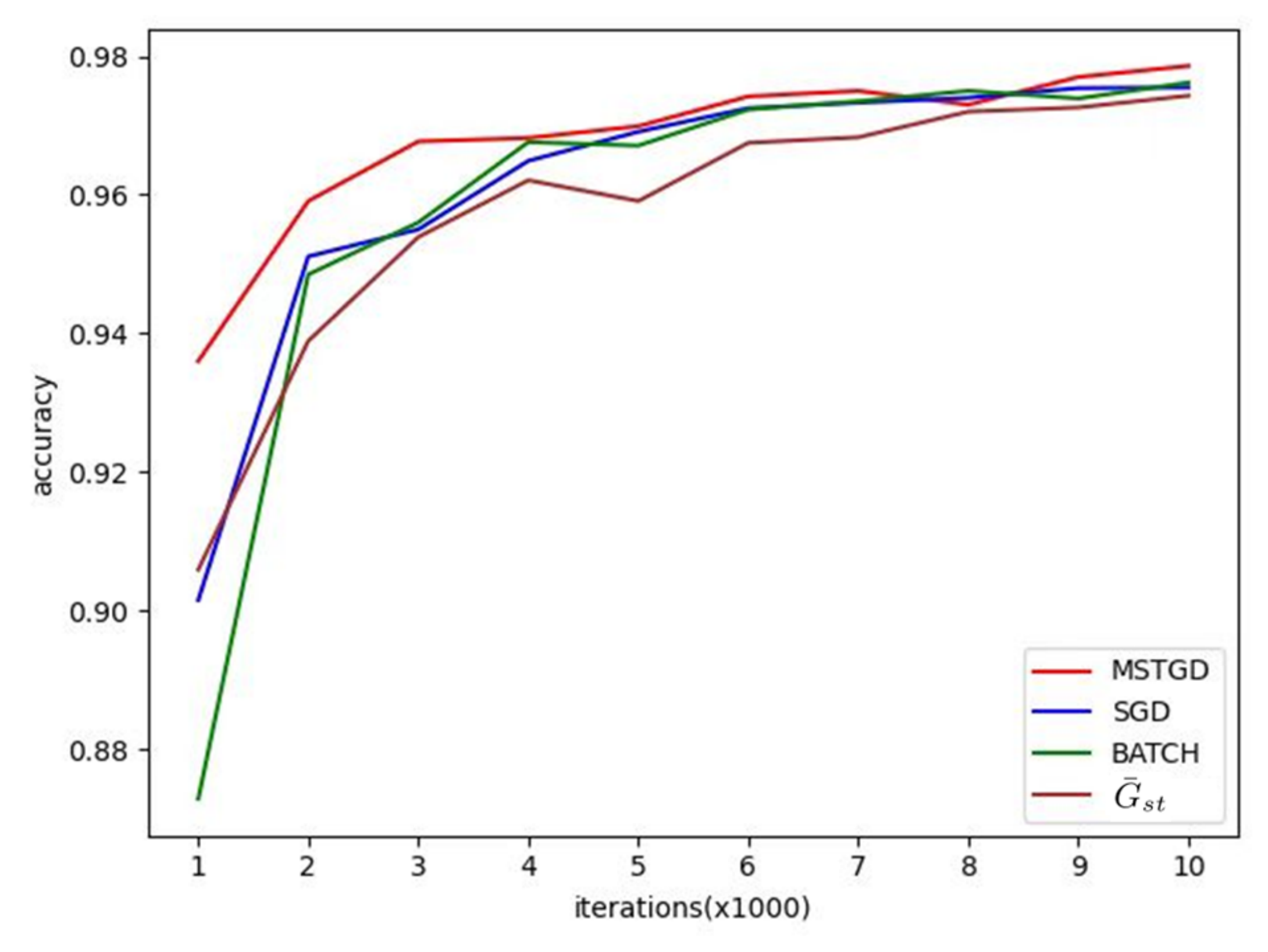}
\caption{\small{(CNN on MNIST)Convergence rate of MSTGD,SGD,Batch,$\bar{G}_{st}$}}
\label{fig:Linearoncnn}
\end{figure}

Figure \ref{fig:Linearoncnn} is the change curve of the average test accuracy of the four algorithms MSTGD, SGD, Batch, $\bar{G}_{st}$ when training the convolutional network for 14 times. It can also be seen that, the accuracy improvement of MSTGD significantly requires fewer iterations than other algorithms. This shows that MSTGD also exhibits a faster convergence rate when training convolutional networks with MNIST data

From the above experimental results, MSTGD indeed exhibits a fast convergence rate, which is consistent with the theoretical results of Theorem \ref{th:linear} on the linear convergence of MSTGD in this paper

\section{Comparison of related works}
The algorithm MSTGD in this paper uses  historical gradient and stratified sampling for variance reduction, so that the algorithm can achieve a linear convergence rate under a constant step size. Section \ref{subsect:linearrate} compares MSTGD with other algorithms such as SAG, SVRG, SAGA which use historical gradient and claim to achieve a linear convergence rate. Section \ref{subsect:sampling} compares MSTGD with other algorithms using different sampling strategies. Section \ref{subsect:otherclassic} compares MSTGD with other classic algorithms using historical gradient strategies.

\subsection{Comparison of Related Algorithms for Variance Reduction Based on Historical Gradients}\label{subsect:linearrate}
SAG, SAGA  use a same storage structure G to record historical gradient information. Each time a random sample i($i\in \{1,2,\cdots,N\}$) is selected,the current gradient $J'_i(W_k)$  is calculated and replaced  with the old gradient value $J'_i(W_{old})$  in the $i^{th}$ component in G. This means that SAG, SAGA both need to save N stochastic gradient vectors, which is a lot of storage overhead in the case of massive training samples. MSTGD only needs to save C (C is the number of categories,$C\ll N$) stochastic gradient vectors.

SAG,SAGA have slightly different variance reduction strategies designed around the historical gradient in G. This difference results in that the gradient update direction of SAG is not an unbiased estimate of the full gradient direction, while the gradient update direction of SAGA and MSTGD are both unbiased estimates of the full gradient direction.

Zhang Tong et al. proposed a progressively stochastic gradient variance reduction algorithm SVRG\cite{Johnson2013}\ and its extension Prox-SVRG\cite{Xiao2014}. The core idea of their algorithm is to calculate and record the full gradient $\nabla J(\widetilde{W})=\frac{1}{N}\sum\limits_{i=1}^N J'_i(\widetilde{W})$ on a reference network $\widetilde{W}$ of the outer loop, and then use this $\nabla J(\widetilde{W})$ to reduce the variance of the stochastic gradient in the inner loop that performs parameter update. Based on this idea,SVRG, Prox-SVRG can achieve effective variance reduction while maintaining unbiasedness, and have been theoretically proved to obtain a linear convergence rate.

The variance reduction strategies of SAG, SAGA, SVRG are similar. They all calculate and save a full gradient $\nabla J=\frac{1}{N}\sum\limits_{i=1}^N J'_i$ in advance, and then use the recorded full gradient $\nabla J$ for variance reduction during the iterative parameter update process. This is essentially different from  p-based variance reduction strategy of MSTGD.
\subsection{Comparison of Related Algorithms for Variance Reduction Based on Sampling Strategy}\label{subsect:sampling}
In order to achieve a better stochastic gradient variance reduction and accelerate algorithm Convergence, Zhang Tong et al. proposed an importance scoring scheme that assigns weight $Pr_i^k$ to sample i in iteration k, and then performs importance sampling according to distribution $Pr_i^k$, replacing the original uniform sampling when selecting a random sample\cite{Zhao2014,Zhao2015}.From their experimental results, compared with uniform sampling, although the importance sampling strategy shows obvious advantages in the decline rate of the objective function value, there is no obvious improvement in the test accuracy of the trained model.

Fartash Faghri et al. studied the gradient distribution of deep learning. They proposed an idea of gradient clustering and stratified sampling according to the categories obtained by the clustering\cite{Fartash2020}. The method developed by Fartash Faghri et al. can find the optimal clustering scheme that minimizes the variance of the stratified mean of gradients. However, the work of Fartash Faghri et al. give neither further information on training the deep network based on the obtained optimal stratified mean of gradients,npor final conclusions about the effect of their variance scheme on the test accuracy of the network.

Different from the above, MSTGD directly uses the category information in the supervision signal Y to perform stratified sampling without any additional knowledge.

\subsection{Comparison of other classical algorithms}\label{subsect:otherclassic}
From the update formula $\upsilon_t=\gamma\upsilon_{t-1}+\eta\nabla_{\theta} J(\theta)$ and $\theta=\theta-\upsilon_t$
of the Momentum optimization\cite{Qian1999},
the gradient direction required for parameter update in the Momentum optimization is obtained
as the weighted sum of the current gradient $\nabla_{\theta} J(\theta)$ and
the historical gradient information stored in $\upsilon_{t-1}$ with weights $\eta$ and $\gamma$.
Here, $\eta$ and $\gamma$ are similar to the parameters $q$ and $p$,
and $\upsilon_{t}$ corresponds to the auxiliary variable $G$ in our model.
However, the unbiasedness of $\upsilon_{t}$ has not been discussed.
The results of this paper show that if $\eta$ and $\gamma$ satisfy the conditions in formula \ref{eq:valuepq} as $q$ and $p$,
the search direction provided by Momentum optimization satisfies unbiasedness.

Xie et al.\cite{Xie2021} propose a Positive-Negative Momentum (PNM) approach. Similar to the traditional Momentum method, their work does not discuss the influence of the choice of Momentum coefficients on unbiasedness, and their work mainly focuses on how to simulate the noise in the stochastic gradient to enhance the generalization ability of the network.

The popular algorithm of Adam \cite{Kingma2015} for training deep models is of gradient update formula
$m_t=\beta_1m_{t-1}+(1-\beta_1)g_t$ and $ \hat{m}_t=\frac{m_t}{1-\beta_1}$.
The parameters $\beta_1$ and $(1-\beta_1)$ therein are respectively equivalent to $p$ and $q$.
According to the results of the current paper, one of the prerequisites for the effectiveness of Adam's coefficients
$\beta_1$ and $(1-\beta_1)$ of $m_{t-1},g_t$ is to ensure that the mean value of the gradients before and after the iterations are equal. Generally, the equal-mean properties of the gradient before and after the iteration are generally not satisfied,
unless additional restrictions are introduced.
Therefore, the author of the Adam algorithm posed a so-called unbiased correction $\hat{m}_t=\frac{m_t}{1-\beta_1}$ on $m_t$.
Obviously, this is an empirical correction formulae, the theoretical basis behind the correction is not fully understood.
In fact, the revised estimator must be a biased estimator, which is contrary to the original intention of the proponent.

Other variance reduction methods\cite{Blatt2007}\ that perform $k$-step averaging
on historical trajectories are equivalent to the method in this paper
with $p=\frac{k-1}{k}$ and $q=\frac{1}{k}$.
However, without an additional strategy to ensure equal gradient mean between different iterations,
the unbiasedness of this approach cannot be satisfied, and the effect of the algorithm will be difficult to guarantee.


\section{Appendix}
In this Appendix, we present the proofs of main conclusions in this paper.
\subsection{Proof of Theorem \ref{the:unbias}}
\begin{proof}
In order to ensure that the starting point $\bar{G}_{mst}^1$ of the sequence $\{\bar{G}_{mst}^k\}$ is
an unbiased estimate of $\bar{G}^{1}$, without loss of generality, we randomly select a sample from each category $j\in \{1,2,\cdots C\}$,
calculate its gradient, fill the $G$ vector and calculate $\bar{G}_{mst}^1$ accordingly.
Now, at the beginning, we have $\bar{G}_{mst}^1=\bar{G}_{st}^1=\sum\limits_{j=1}^C w_jG_j^1$.

Taking the expectation of $\bar{G}_{mst}^1$,
we have
$E(\bar{G}_{mst}^1)=E(\bar{G}_{st}^1)=E(\sum\limits_{j=1}^C w_jG_j^1)=\sum\limits_{j=1}^C\textit{w}_jE(G_j^1)=\bar{G}^1$.
Thus the unbiasedness of the starting point of the gradient sequence is established.

Similarly, for the $k$-th iteration, we have
\begin{equation}
\begin{array}{rl}
E(\bar{G}_{mst}^k)=&E(\sum\limits_{j=1}^C\textit{w}_jG_j^k)=\sum\limits_{j=1}^C\textit{w}_jE(G_j^k)\\
=&\sum\limits_{j=1}^C\textit{w}_jE[p_j^kG_j^{k-1}+q_j^kg(W_k,\xi_j^k)]\\
=&\sum\limits_{j=1}^C\textit{w}_j[p_j^kE(G_j^{k-1})+q_j^kE(g(W_k,\xi_j^k))]\\
=&\sum\limits_{j=1}^C\textit{w}_j[p_j^kE_j^{k-1}+q_j^kE_j^k]\\
=&\sum\limits_{j=1}^C\textit{w}_j[(1-q_j^k)E_j^k+q_j^kE_j^k]\\
=&\sum\limits_{j=1}^C\textit{w}_jE_j^k=\bar{G}^k,
\nonumber
\end{array}
\end{equation}
and this concludes the proof.
\end{proof}
\subsection{Proof of Lemma \ref{Lem:SPVar}}\label{appendix:valuepq}
\begin{proof}
We already know that
\ $\bar{G}_{mst}^k=\sum\limits_j^C\textit{w}_jG_j^k=\sum\limits_j^C[\textit{w}_j(p_j^kG_j^{k-1}+q_j^kg(W_k,\xi_j^k))]$\ ,
and the stochastic gradient $g(W_k,\xi_j^k))$ generated by independent sample and the historical gradient $G_j^{k-1}$ in the auxiliary storage are independent of each other. Therefore, the variance of $\bar{G}_{mst}^k$ is:
\begin{equation}\label{eq:var}
\begin{array}{rl}
V(\bar{G}_{mst}^k)=&\sum\limits_j^C[\textit{w}_j^2((p_j^k)^2V_j^{k-1}+(q_j^k)^2V(g(W^k,\xi_j^k)))]\\
=&\sum\limits_j^C[\textit{w}_j^2(\underbrace{(p_j^k)^2V_j^{k-1}+(q_j^k)^2V_j^k}_{z})]
\end{array}
\end{equation}
Since $\bar{G}_{mst}^k$ is an unbiased estimator, substituting the unbiased condition
$\frac{p_j}{1-q_j }=\frac{E_j^k}{E_j^{k-1}}$
in Theorem \ref{the:unbias} into the variable z in formulae (\ref{eq:var})  leads to:
\begin{equation}\label{eq:var2}
\begin{array}{rl}
z=&(1-q_j^k)^2(\frac{E_j^k}{E_j^{k-1}})^2V_j^{k-1}+(q_j^k)^2V_j^k.
\end{array}
\end{equation}
When \ $p_j^k,q_j^k$\ take the values according to formulae (\ref{eq:valuepq}),
$z=\frac{(E_j^k)^2V_j^{k-1}\cdot V_j^{k}}{(E_j^k)^2V_j^{k-1}+(E_j^{k-1})^2V_j^{k}}$.
Substituting it back into Equation (\ref{eq:var})
we get the final variance expression in the form of Equation (\ref{eq:Vmin}).
We conclude the proof.
\end{proof}
\subsection{Proof of Corollary \ref{cor:desigeneffect}}
\begin{proof}
We deduce (1) from the equivalent transformation of Equation (\ref{eq:Vmin}).
Setting $\rho_j=\frac{(E_j^k/E_j^{k-1})^2}{(E_j^k/E_j^{k-1})^2+V_j^k/V_j^{k-1}}<1$,
and taking $\rho=\max\limits_j\rho_j,j\in\{1,\cdots,C\}$, we have:
\begin{equation}
\begin{array}{rl}
V_{sp}(\bar{G}_{mst}^k)=&\sum\limits_j^C[\textit{w}_j^2\frac{(E_j^k)^2V_j^{k-1}\cdot V_j^{k}}{(E_j^k)^2V_j^{k-1}+(E_j^{k-1})^2V_j^{k}}]\\
=&\sum\limits_j^C[\textit{w}_j^2\frac{(E_j^k/E_j^{k-1})^2\cdot V_j^{k}}{(E_j^k/E_j^{k-1})^2+V_j^{k}/V_j^{k-1}}]\\
=&\sum\limits_j^C[\textit{w}_j^2\rho_jV_j^k]\\
\leq&\rho\sum\limits_j^C[\textit{w}_j^2V_j^k]=\rho V(\bar{G}_{st}^k)\\
<&V(\bar{G}_{st}^k) \nonumber
\end{array}
\end{equation}
Therefore, property (1) holds.

Next we prove (2) by induction.

Let $$p=\max\limits_{\substack{j\in\{1,\cdots,C\},\\ k\in\{1,2,\cdots,\}}}p_j^k,
\text{ and } q=\max\limits_{\substack{j\in\{1,\cdots,C\},\\ k\in\{1,2,\cdots,\}}}q_j^k.$$
When $k=1$, $\bar{G}_{mst}^1=\bar{G}_{st}^1$ by Theorem (\ref{the:unbias}). And according to formulae (\ref{eq:weightupdate}),
we have
\begin{equation}\label{eq:bargmst2}
\begin{array}{rl}
\bar{G}_{mst}^2=&\sum\limits_{j=1}^C\textit{w}_j(p_j^2\cdot G_{j}^{1}+q_j^2\cdot g(W_2,\xi_j^2))\\
=&\sum\limits_{j=1}^C\textit{w}_j(p_j^2\cdot G_{j}^{1})+\sum\limits_{j=1}^C\textit{w}_j(q_j^2\cdot g(W_2,\xi_j^2))\\
\leq &p\cdot\sum\limits_{j=1}^C\textit{w}_jG_{j}^{1}+q\cdot\sum\limits_{j=1}^C\textit{w}_jg(W_2,\xi_j^2)\\
=&p\cdot \bar{G}_{mst}^1+q\cdot \bar{G}_{st}^2.
\end{array}
\end{equation}

Let $p_j^2$ and $q_j^2$ be given by formulae (\ref{eq:valuepq}),we get
\begin{equation}\label{eq:bargmst3}
\begin{array}{rl}
V_{sp}(\bar{G}_{mst}^2)=&V[\sum\limits_{j=1}^C\textit{w}_j(p_j^2\cdot G_{j}^{1})]+V[\sum\limits_{j=1}^C\textit{w}_j(q_j^2\cdot g(W_2,\xi_j^2))]\\
\leq &p^2\cdot V[\sum\limits_{j=1}^C\textit{w}_jG_{j}^{1}]+q^2\cdot V[\sum\limits_{j=1}^C\textit{w}_jg(W_2,\xi_j^2)]\\
=&p^2\cdot V(\bar{G}_{mst}^1)+q^2\cdot V(\bar{G}_{st}^2),
\end{array}
\end{equation}
which means (2) holds when $k=1,t=1$.

Similar to (\ref{eq:bargmst2}) and (\ref{eq:bargmst3}) we can verify that $V(\bar{G}_{mst}^{k+1})\leq p^2V(\bar{G}_{mst}^k)+q^2V(\bar{G}_{st}^{k+1})$,
that is, (2) is valid when $t=1$ and $k$ takes any value.


Now assume that when $t=n$ and $k$ takes an arbitrary value, (2) holds.
Below we show that when $t=n+1$ and $k$ remains unchanged, (2) still holds.
\begin{equation}\label{eq:bargmstk}
\begin{array}{rl}
V_{sp}(\bar{G}_{mst}^{k+n+1})\leq&p^2\cdot V(\bar{G}_{mst}^{k+n})+q^2\cdot V(\bar{G}_{st}^{k+n+1})\\
\leq&p^2\cdot [p^{2n}V(\bar{G}_{mst}^{k})+\sum\limits_{i=1}^np^{2(n-i)}q^2V(\bar{G}_{st}^{k+i})]+q^2\cdot V(\bar{G}_{st}^{k+n+1})\\
=&p^{2(n+1)}V(\bar{G}_{mst}^{k})+p^2\sum\limits_{i=1}^np^{2(n-i)}q^2V(\bar{G}_{st}^{k+i})+q^2\cdot V(\bar{G}_{st}^{k+n+1})\\
=&p^{2(n+1)}V(\bar{G}_{mst}^{k})+\sum\limits_{i=1}^{n+1}p^{2(n+1-i)}q^2V(\bar{G}_{st}^{k+i}).
\end{array}
\end{equation}

In summary, (2) holds for any $k$ and $t$, and the proof is complete.
\end{proof}
\subsection{Proof of Lemma \ref{Lem:linearslowthanexp}}
\begin{proof}
Divide both sides of inequality (\ref{Lem:linearslowthanexp}) by $\eta^{2t}$ and multiply by $\gamma^{2k_0}$ to get a new equivalent inequality
\begin{equation}\label{ieq:rhogamma}
(1+\frac{1}{1-\eta})\gamma^{2k_0}\leq (\frac{\gamma}{\eta})^{2k}.
\end{equation}
Let $0<\eta<\gamma<1$, then $\frac{\gamma}{\eta}>1$ . Given the values of\ $\gamma,\eta,k_0$\ ,the left side of inequality (\ref{ieq:rhogamma}) is a constant, and the right side is a function about the exponential growth of 2k.In the case of sufficiently large k, inequality (\ref{ieq:rhogamma}) is obviously established, so inequality (\ref{Lem:linearslowthanexp}) is also established.

\end{proof}
\subsection{Proof of Theorem \ref{The:p-less-one}}\label{appendix:plessone}
\begin{proof}
Let $p=\max\limits_{\substack{j\in\{1,\cdots,C\},\\ k\in\{1,2,\cdots,\}}}p_j^k$. In the following two case, $p\leq 1$ holds.

When $E_j^k=E_j^{k-1}$, according to Equation (\ref{eq:valuepq})  and the fact $(E_j^k)^2V_j^{k-1}\geq 0$, we have $p_j^k\leq 1$.So $p\leq 1$.

Further let $\frac{0}{0}=1$. When $E_j^k=E_j^{k-1}=0$, according to (\ref{The:valuepq}) and the fact $V_j^{k-1}\geq 0$, we have $p_j^k=\frac{\frac{E_j^k}{E_j^{k-1}}V_j^k}{(\frac{E_j^k}{E_j^{k-1}})^2V_j^{k-1}+V_j^k}=\frac{V_j^k}{V_j^{k-1}+V_j^k}\leq 1$. Also $p\leq 1$ holds.
p

we know
\begin{equation}
V_{sp}(\bar{G}_{mst})=\begin{tiny}\left(\begin{array}{cccccc}\sigma_1^2&&&&&\\&\sigma_2^2&&&&\\&&\bullet&&&\\&&&\bullet&&\\&&&&\bullet&\\&&&&&\sigma_{\wp}^2\end{array}\right)\end{tiny},
V(\bar{G}_{st}^i)=\begin{tiny}\left(\begin{array}{cccccc}\Sigma_{1i}^2&&&&&\\&\Sigma_{2i}^2&&&&\\&&\bullet&&&\\&&&\bullet&&\\&&&&\bullet&\\&&&&&\Sigma_{\wp i}^2\end{array}\right)\end{tiny}\nonumber
\end{equation}


Let\ $\rho=max(p,q),M=max\{\sigma_1^2,\sigma_2^2,\cdots,\sigma_{\wp}^2,\Sigma_{1i}^2,\Sigma_{2i}^2,\cdots,\Sigma_{\wp i}^2\forall i\}$\ . When $p\leq 1$, according to property (2) in Corollary(\ref{cor:desigeneffect}), we have
\begin{equation}\label{ieq:gamma}
\begin{array}{rl}
V_{sp}(\bar{G}_{mst}^{k})\leq & p^{2k}V(\bar{G}_{mst})+\sum\limits_{i=1}^{t}p^{2(k-i)}q^2V(\bar{G}_{st}^{i})\\
\leq &p^{2k}M\cdot I+\sum\limits_{i=1}^{t}p^{2(k-i)}q^2M\cdot I\\
= &p^{2k}M\cdot I+\frac{p^{2(k-1)}}{1-p^2}q^2M\cdot I\\
\leq &(1+\frac{1}{1-p^2})\rho^{2k}M\cdot I\\
\leq &(1+\frac{1}{1-p})\rho^{2k}M\cdot I\\
\leq &(1+\frac{1}{1-\rho})\rho^{2k}M\cdot I\\
\leq &\gamma^{2(k-k_0)}M\cdot I.\\
\end{array}
\end{equation}
\end{proof}
The equation of step 3 in formulae (\ref{ieq:gamma}) is established because the proportional sequence in the previous formula is summed, and the last inequality is established because of Lemma (\ref{Lem:linearslowthanexp}).
The proof is concluded.
\subsection{Proof of Theorem \ref{The:VofbarGmst}}\label{VofbarGmst}
\begin{proof}
Let $\sigma_i^2=E[\bar{G}_{mst}^k(w_i)]^2-[E(\bar{G}_{mst}^k(w_i))]^2$. According to the definitions of $V_{sp}(\bar{G}_{mst}^k),Var(\bar{G}_{mst}^k)$,we know
\begin{equation}
V_{sp}(\bar{G}_{mst})=\begin{tiny}\left(\begin{array}{cccccc}\sigma_1^2&&&&&\\&\sigma_2^2&&&&\\&&\bullet&&&\\&&&\bullet&&\\&&&&\bullet&\\&&&&&\sigma_{\wp}^2\end{array}\right)\end{tiny},
Var(\bar{G}_{mst}^k)=\sum\limits_{i=1}^{\wp}\sigma_i^2,\nonumber
\end{equation}
i.e. $Var(\bar{G}_{mst}^k)$\ is the trace of $V_{sp}(\bar{G}_{mst})$. Applying Theorem (\ref{The:p-less-one}), we have
\begin{equation}
\begin{array}{rl}
Var(\bar{G}_{mst}^k)=&\sum\limits_{i=1}^{\wp}\sigma_i^2\\
=&tr(V_{sp}(\bar{G}_{mst}))\\
\leq& \gamma^{2(k-k_0)}\wp M\\
=&\gamma^{2(k-k_0)}\Phi .
\end{array}
\end{equation}

This conclude the proof.
\end{proof}
\subsection{Proof of Inequality (\ref{ieq:strongconvex})}\label{proof:strongconvex}
\begin{proof}
For a quadratic model
\begin{center}
$f(\bar{W})=J(W)+\nabla J(W)^T(\bar{W}-W)+\frac{1}{2}||\bar{W}-W||_2^2$
\end{center}

with respect to $W\in \Re^{\wp}$, its unique minima is $\bar{W}_*=W-\frac{1}{c}\nabla J(W)$ and minimal value $f(\bar{W}_*)=J(W)-\frac{1}{2c}||\nabla J(W)||_2^2$.

Taking $W'=W_*$ for (\ref{eq:12}) will have $J_*\leq J(W)+\nabla J(W)^T(W_*-W)+\frac{1}{2}c||W_*-W||_2^2\leq J(W)-\frac{1}{2c}||\nabla J(W)||_2^2$, so inequality (\ref{ieq:strongconvex}) holds.
\end{proof}
\subsection{Proof of Theorem \ref{th:cvi}}\label{appendix:cvi}
\begin{proof}
According to assumption $A_1$, we have
\begin{equation}
\begin{array}{rl}
J(W_{k+1})=& J(W_k)+\int_0^1\frac{\partial J(W_k+t(W_{k+1}-W_k))}{\partial t}dt\\
=& J(W_k)+\int_0^1 \nabla J(W_k+t(W_{k+1}-W_k))^T(W_{k+1}-W_k)dt\\
=& J(W_k)+\nabla J(W_k)^T(W_{k+1}-W_k)+\\
&\  \int_0^1 [\nabla J(W_k+t(W_{k+1}-W_k))-\nabla J(W_k)]^T(W_{k+1}-W_k)dt\\
\leq & J(W_k)+\nabla J(W_k)^T(W_{k+1}-W_k)+\int_0^1L||t(W_{k+1}-W_k)||_2||W_{k+1}-W_k||_2dt\\
= &J(W_k)+\nabla J(W_k)^T(W_{k+1}-W_k)+\frac{1}{2}||W_{k+1}-W_k||_2^2.
\end{array} \nonumber
\end{equation}

This leads to
\begin{equation}\label{ieq:JW-JWPrime}
\begin{array}{rl}
J(W_{k+1})-J(W_k)\leq &\nabla J(W_k)^T(W_{k+1}-W_k)+\frac{1}{2}L||W_{k+1}-W_k||_2^2.
\end{array}
\end{equation}

Substituting formulae (\ref{eq:sgd-likeupdate})  into the above inequality and taking expectation on both sides, we have
\begin{equation}\label{eq:17}
\small
\begin{array}{rl}
E[J(W^{k+1})]-J(W_{k})\leq & -h_k\nabla J_N(W_{k})^TE[\bar{g}]+\frac{1}{2}LE[||\bar{g}||^2_2]h_k^2 \\
=&-h_k\nabla J_N(W_{k})^TE[\bar{g}]+\frac{1}{2}L(s_k^2+ ||E[\bar{g}]||^2_2)h_k^2\\
=&-h_k\nabla J_N(W_{k})^T\nabla J_N(W_{k})+\frac{1}{2}L(s_k^2+ \nabla J_N(W_{k})^2_2)h_k^2\\
=&(\frac{h_k^2L}{2}-h_k) ||\nabla J_N(W_{k})||^2_2+\frac{h_k^2Ls_k^2}{2}.
\end{array}
\end{equation}
The first equation in the above formula holds because of the known conclusion that $s_k^2=Var(\bar{g})=E[||\bar{g}||_2^2]-||E(\bar{g})||_2^2$,and the second equation holds because the expectation of the sample mean is equal to the population mean, that is, $E[\bar{g}]=\nabla J_N(W_{k})$.

With the condition $h_k<\frac{2}{L}$ and the fact (\ref{ieq:strongconvex}), formula \eqref{eq:17} will be changed as
\begin{equation}
\begin{array}{rl}
E[J(W^{k+1})]-J(W^{k})\leq & (\frac{h_k^2L}{2}-h_k) ||\nabla J_N(W^{k})||^2_2+\frac{h_k^2Ls_k^2}{2}\\
\leq& (\frac{h_kL}{2}-1)2h_kc(J(W^{k})-J^*)+\frac{h_k^2Ls_k^2}{2}. \nonumber
\end{array}
\end{equation}

Subtracting $J^*$ on both sides of the above inequality, taking expectation and reordering it, we have
\begin{equation}
\begin{array}{l}
E[J(W^{k+1})-J^*]\leq (h_k^2c L-2h_kc+1)E[J(W^{k})-J^*]+\frac{h_k^2Ls_k^2}{2}. \nonumber
\end{array}
\end{equation}

Let $e_k=E[J(W^{k})-J^*],\Lambda_k=\frac{h_kLs_k^2}{2c(2-h_kL)}(c\neq 0,h_kL\neq 2), \rho_i=(h_i^2c L-2h_ic+1)<1,\rho=\max\limits_{i=1,\cdots,k}\{\rho_i\}$, then
\begin{equation}\label{eq:contraction}
\begin{array}{rl}
E[J(W^{k+1})]-J_*-\Lambda_{k}\leq & (h_k^2c L-2h_kc+1)(E[J(W^{k})-J_*]-\Lambda_{k}),\\
e_{k+1}-\Lambda_{k}\leq & \rho_k(e_k-\Lambda_{k}).
\end{array}
\end{equation}

In addition, let\ $\sigma^2_{min}=\min\{\sigma^2_1,\sigma^2_2,\cdots,\sigma^2_k,\cdots,\},\sigma^2_{max}=\max\{\sigma^2_1,\sigma^2_2,\cdots,\sigma^2_k,\cdots,\}$\ ,we can carefully select the ratio of sample size\ $\frac{n_{k+1}}{n_k}$\ in adjacent iterations such that it satisfies the following inequality
\begin{equation}\label{eq:ratioofsamplesize}
\begin{array}{l}
\frac{n_{k+1}}{n_k}\geq r^2\frac{h_{k+1}}{h_k}\frac{2-h_kL}{2-h_{k+1}L},
\end{array}
\end{equation}
where\ $r^2=\frac{\sigma^2_{min}}{\sigma^2_{max}}\leq 1$\ . When\ $h_k\rightarrow 0,\forall k=1,2,\cdots$\ , formulae \eqref{eq:ratioofsamplesize} can be simplified as\ $\frac{n_{k+1}}{n_k}\geq r^2$\ , this is a  weak and easily satisfied condition.

We can verify formulae \eqref{eq:ratioofsamplesize} leads to $\Lambda_{k-1}-\Lambda_{k}>0$ .

Applying the inequality $e_{k+1}-\Lambda_{k}\leq \rho_k(e_k-\Lambda_{k})$ recursively, with the guarantee of formulae \eqref{eq:ratioofsamplesize},  formulae \eqref{eq:contraction} will be changed as
\begin{equation}\label{eq:contraction2}
\begin{array}{l}
e_{k+1}-\Lambda_{k}\\
\leq  \rho_k(e_k-\Lambda_{k})= \rho_k(e_k-\Lambda_{k-1}+\Lambda_{k-1}-\Lambda_{k})\\
\leq  \rho_k[\rho_{k-1}(e_{k-1}-\Lambda_{k-1})+\Lambda_{k-1}-\Lambda_{k}]=\rho_k\rho_{k-1}(e_{k-1}-\Lambda_{k-1})+\rho_k(\Lambda_{k-1}-\Lambda_{k})\\
\leq \rho_k\rho_{k-1}\rho_{k-2}(e_{k-2}-\Lambda_{k-2})+\rho_k\rho_{k-1}(\Lambda_{k-2}-\Lambda_{k-1})+\rho_k(\Lambda_{k-1}-\Lambda_{k})\\
\ldots\ldots\ldots\\
\leq  (e_{1}-\Lambda_{1})\prod\limits_{i=1}^k\rho_{i}+(\Lambda_{1}-\Lambda_{2})\prod\limits_{i=2}^k\rho_{i}+(\Lambda_{2}-\Lambda_{3})\prod\limits_{i=3}^k\rho_{i}+\cdots+\\
\qquad  \qquad \qquad \qquad \qquad \qquad \qquad \qquad(\Lambda_{k-2}-\Lambda_{k-1})\prod\limits_{i=k-1}^k\rho_{i}+(\Lambda_{k-1}-\Lambda_{k})\rho_{k}\\
\leq (e_{1}-\Lambda_{1})\prod\limits_{i=1}^k\rho_{i}+\sum\limits_{j=2}^{k}[(\Lambda_{j-1}-\Lambda_{j})\prod\limits_{i=j}^k\rho_{i}].
\end{array}
\end{equation}

Formulae \eqref{eq:contraction2} shows that the difference of adjacent iterations $\Lambda_{j-1}-\Lambda_{j}$ decays at a rate of $\prod\limits_{i=j}^k\rho_{i}$ during the iteration.\\

Considering an extreme case, the difference of adjacent iterations $\Lambda_{j-1}-\Lambda_{j}$ will cease decaying by removing the coefficient $\prod\limits_{i=j}^k\rho_{i}$ from the term $(\Lambda_{j-1}-\Lambda_{j})\prod\limits_{i=j}^k\rho_{i}$, and let $\rho=\max\limits_i\rho_i$,we have
\begin{equation}\label{eq:contraction3}
\begin{array}{rl}
e_{k+1}\leq&  (e_{1}-\Lambda_{1})\prod\limits_{i=1}^k\rho_{i}+(\Lambda_{1}-\Lambda_{2})+\cdots+(\Lambda_{k-1}-\Lambda_{k}) +\Lambda_{k}\\
<&\Lambda_{1}+(e_{1}-\Lambda_{1})\prod\limits_{i=1}^k\rho_{i}\\
\leq & \Lambda_{1}+\rho^k(e_{1}-\Lambda_{1}).
\end{array}
\end{equation}

Taking $e_k=E[J(W^{k})-J_*]$ back into above inequality, the final result of Theorem \ref{th:cvi} is derived.
\end{proof}

\subsection{Proof of Theorem \ref{th:linear}}\label{appendix:linear}
\begin{proof}
The step size in this proof is a constant step size $\bar{h}$.
Substituting $W_{k+1}=W_k-\bar{h}\bar{G}^k_{mst}$(formulae \ref{eq:weightupdate}) into inequality (\ref{ieq:JW-JWPrime}),we get
\begin{equation}
\begin{array}{rl}
J(W_{k+1})-J(W_k)\leq &\nabla J(W_k)^T(W_{k+1}-W_k)+\frac{1}{2}L||W_{k+1}-W_k||_2^2\\
\leq & -\bar{h}\nabla J(W_k)^T\bar{G}_{mst}^k+\frac{1}{2}\bar{h}^2L||\bar{G}_{mst}^k||_2^2.
\end{array} \nonumber
\end{equation}

Since $E(\bar{G}_{mst}^k)=\bar{G}^k=\nabla J(W_k)$, substituting (\ref{eq:weightupdate}) into the above inequality, and taking the expectation on both sides, the above inequality can be transformed into the following form
\begin{equation}
\begin{array}{rl}
E(J(W_{k+1})-J(W_k)\leq &-\bar{h}\nabla J(W_k)^TE(\bar{G}_{mst}^k)+\frac{1}{2}\bar{h}^2LE[||\bar{G}_{mst}^k||_2^2]\\
= & -\bar{h}||\nabla J(W_k)||_2^2+\frac{1}{2}\bar{h}^2LE[||\bar{G}_{mst}^k||_2^2].
\end{array} \nonumber
\end{equation}

Assuming $\mu_G\geq\mu\geq 0$, such that for all $k\in N$\ ,$\nabla J(W_k)^TE(\bar{G}_{mst}^k)\geq\mu||\nabla J(W_k)||_2^2$, and $||E(\bar{G}_{mst}^k)||_2\leq\mu_G||\nabla J(W_k)||_2$, the above inequality can be further transformed into the following form
\begin{equation}
\begin{array}{rl}
E(J(W_{k+1})-J(W_k)\leq &-\mu \bar{h}||\nabla J(W_k)||_2^2+\frac{1}{2}\bar{h}^2LE[||\bar{G}_{mst}^k||_2^2]\\
\end{array} \nonumber
\end{equation}
Let $Var(\bar{G}_{mst}^k):=E(||\bar{G}_{mst}^k||_2^2)-||E(\bar{G}_{mst}^k)||_2^2$\ ,Applying \ $||E(\bar{G}_{mst}^k)||_2\leq \mu_G||\nabla J(W_k)||_2,Var(\bar{G}_{mst}^k)\leq \gamma^{2(k-k_0)}\Phi$(Theorem \ref{The:VofbarGmst}\ ),$0<\bar{h}\leq min\{\frac{\mu}{L\mu_G^2},\frac{1}{c\mu}\}$\ , we have
\begin{equation}
\begin{array}{rl}
E(J(W_{k+1}))-J(W_k)\leq &-\mu \bar{h}||\nabla J(W_k)||_2^2+\frac{1}{2}\bar{h}^2L(\mu_G^2||\nabla J(W_k)||_2^2+\Phi\gamma^{2(k-k_0)})\\
\leq &-(\mu-\frac{1}{2}L\mu_G^2)\bar{h}||\nabla J(W_k)||_2^2+\frac{1}{2}\bar{h}^2L\Phi\gamma^{2(k-k_0)}\\
\leq &-\frac{1}{2}\mu\bar{h}||\nabla J(W_k)||_2^2+\frac{1}{2}\bar{h}^2L\Phi\gamma^{2(k-k_0)}.
\end{array} \nonumber
\end{equation}
Under the strong convex assumption and the fact (\ref{ieq:strongconvex}), we further have
\begin{equation}
\begin{array}{rl}
E(J(W_{k+1}))-J(W_k)\leq &-\frac{1}{2}\mu\bar{h}||\nabla J(W_k)||_2^2+\frac{1}{2}\bar{h}^2L\Phi\gamma^{2(k-k_0)}\\
\leq & -\bar{h}c\mu(J(W_k)-J_*)+\frac{1}{2}\bar{h}^2L\Phi\gamma^{2(k-k_0)}.
\end{array} \nonumber
\end{equation}
Subtracting $J_*$ and adding $J(W_k)$ from both sides of the above inequality, rearranging and taking expectation on both sides, we get
\begin{equation}\label{ieq:gapofcost}
\begin{array}{rl}
E(J(W_{k+1}))-J_*\leq &(1-\bar{h}c\mu)E[J(W_k)-J_*]+\frac{1}{2}\bar{h}^2L\Phi\gamma^{2(k-k_0)}.
\end{array}
\end{equation}
Let
\begin{equation}
\begin{array}{l}
\Omega:=max\{\frac{\bar{h}L\Phi}{c\mu},J(W_{k_0})-J_*\}\\
\lambda:=max\{1-\frac{\bar{h}c\mu}{2},\gamma\}<1
\end{array} \nonumber
\end{equation}
then (\ref{ieq:gapofcost}) can be reformed as
\begin{equation}
\begin{array}{rl}
E[J(W_{k})-J_*]\leq &\Omega \lambda^{2(k-k_0)},
\end{array} \nonumber
\end{equation}
which is what we expected as (\ref{The:linarconvergence}).
The following is an inductive proof for the above inequality.
When $k=k_0$ , we have
\begin{equation}
\begin{array}{rl}
E(J(W_{k_0})-J_*)\leq & \Omega\\
=&\Omega \lambda^{2(k_0-k_0)}.
\end{array} \nonumber
\end{equation}
This shows that inequality (\ref{The:linarconvergence}) holds for $k=k_0$.

Now suppose that when $k\geq k_0$, inequality (\ref{The:linarconvergence}) holds, the following  process
\begin{equation}
\begin{array}{rl}
E[J(W_{k+1})-J_*]\leq &(1-\bar{h}c\mu)E[J(W_k)-J_*]+\frac{1}{2}\bar{h}^2L\Phi\gamma^{2(k-k_0)}\\
\leq &(1-\bar{h}c\mu)\Omega\lambda^{2(k-k_0)}+\frac{1}{2}\bar{h}^2L\Phi\gamma^{2(k-k_0)}\\
=& \Omega\lambda^{2(k-k_0)}(1-\bar{h}c\mu+\frac{\bar{h}^2L\Phi}{2\Omega}(\frac{\gamma}{\lambda})^{2(k-k_0)})\\
\leq & \Omega\lambda^{2(k-k_0)}(1-\bar{h}c\mu+\frac{\bar{h}^2L\Phi}{2\frac{\bar{h}L\Phi}{c\mu}}(\frac{\gamma}{\lambda})^{2(k-k_0)})\\
=&  \Omega\lambda^{2(k-k_0)}(1-\bar{h}c\mu+\frac{\bar{h}c\mu}{2}(\frac{\gamma}{\lambda})^{2(k-k_0)})\\
\leq & \Omega\lambda^{2(k-k_0)}(1-\bar{h}c\mu+\frac{\bar{h}c\mu}{2}\frac{\bar{h}c\mu}{2})\\
=& \Omega\lambda^{2(k-k_0)}(1-\frac{\bar{h}c\mu}{2})^2\\
\leq & \Omega\lambda^{2(k+1-k_0)},
\end{array} \nonumber
\end{equation}
shows that inequality (\ref{The:linarconvergence})  also holds for $k+1$. The proof is concluded.p
\end{proof}

\end{document}